\documentclass[11pt,a4paper]{article}

\usepackage[utf8]{inputenc}
\usepackage[T2A,T1]{fontenc}
\usepackage[ukrainian,english]{babel}
\usepackage{amsmath,amssymb}
\usepackage{graphicx}
\usepackage{booktabs,multirow,array,tabularx}
\usepackage{hyperref}
\usepackage{natbib}
\usepackage[margin=1in]{geometry}
\usepackage{microtype}
\usepackage{enumitem}
\usepackage{caption,subcaption}
\usepackage{float}
\usepackage{xcolor}
\usepackage{underscore}
\usepackage{tikz}
\graphicspath{{figures/}{./}}
\usepackage{pgfplots}
\pgfplotsset{compat=1.18}
\usepgfplotslibrary{colorbrewer}
\usepackage{rotating}

\newcolumntype{R}[1]{>{\raggedleft\arraybackslash}p{#1}}
\newcommand{\cmark}{\textcolor{green!60!black}{\checkmark}}
\newcommand{\xmark}{\textcolor{red!60!black}{--}}


\newcommand{\temporalnpre}{4,908}\newcommand{\temporalnpost}{1,099}

\hypersetup{
    colorlinks=true,
    linkcolor=blue!70!black,
    citecolor=blue!70!black,
    urlcolor=blue!70!black,
}

\title{Multi-Legal-Bench: When the Answer Is in the Input.\\Label Leakage in Legal Benchmarks Built from Court Registries}

\author{
  Volodymyr Ovcharov \\
  LawRider Ltd \\
  \texttt{vladimir@lawrider.uk} \\
  \url{https://lawrider.uk}
}

\date{}

\begin{document}
\maketitle

\begin{abstract}
Court registries publish millions of decisions with structured metadata, which makes them an attractive source of labelled legal benchmarks: the court type, the form of the decision and its subject area come for free. We show that this convenience has a cost that registry-derived benchmarks rarely measure. We build \textbf{Multi-Legal-Bench}, which evaluates identical tasks on native court decisions from five national registries (France, the Netherlands, Poland, the Czech Republic and Lithuania), extending the Ukrainian UA-Legal-Bench, and audit what its cells actually measure. A keyword scan that uses no model reaches 96\% on Dutch judgment-form classification against a 50\% majority baseline, 84\% on French court-type classification against 33\%, and 66\% on Polish judgment-form classification, where all nine models add at most seven points over it. Replacing the label names in the input by a mask brings the scan to the majority floor in all six affected cells, and eight of nine models lose accuracy significantly in most of them (39 of 54 model--cell pairs), by up to 48 points (Dutch judgment form: 100\% to 52\%); only Claude Sonnet~5 is essentially unaffected. Paired McNemar tests with Holm correction find a leader that beats every other model in only two of twelve cells, each a different model; but once the label names are masked, the same model leads both, and the spread between models widens in every masked cell. Part of the apparent parity between models is produced by the leakage itself. The audit also exposed a scoring defect in our own earlier release (answers written with diacritics were scored as wrong, understating Czech judgment-form accuracy by up to 90 points), which we correct and document. We recommend that every benchmark built from court registries report a no-model baseline and a masked control per cell. All data, prompts, predictions and scoring code are released.

\noindent\textbf{Data:} \url{https://huggingface.co/datasets/overthelex/multi-legal-bench}
\end{abstract}

\noindent\textbf{Keywords:} legal NLP, benchmark validity, label leakage, shortcut learning, multilingual evaluation, court decisions

\section{Introduction}
\label{sec:intro}

Legal NLP benchmarks increasingly take their labels from the metadata that courts publish with their decisions. The appeal is obvious. National registries hold millions of decisions, and each already carries a court type, a decision form, an outcome or a subject area, assigned by the court or the registry rather than by an annotator. A benchmark that maps a task onto such a field inherits its scale and its authority.

It also inherits the document. A decision of the French Cour de cassation opens with the words ``Cour de cassation''; a Polish judgment is headed \emph{WYROK}; a Dutch opinion of the Procurator General is published under a metadata line that reads \emph{Conclusie}. When the label a task asks for is printed in the text the model is shown, high accuracy measures the ability to find it, not the legal understanding the task name suggests. This is the legal-domain form of a problem that NLP has met before, in annotation artifacts that let hypothesis-only models solve inference~\citep{gururangan2018annotation,poliak2018hypothesis} and in shortcut learning more generally~\citep{geirhos2020shortcut,mccoy2019right}, and we recently documented a related failure in legal multiple-choice exams, where answer options alone are enough~\citep{ovcharov2026optionleakage}. Registry-derived benchmarks rarely test for it.

We make that test the centre of this paper. \textbf{Multi-Legal-Bench} evaluates identical tasks on native decisions from national registries (Table~\ref{tab:coverage}); this paper evaluates the five non-Ukrainian ones, the Ukrainian cells being those of the companion UA-Legal-Bench~\citep{ovcharov2025ualegalbench}. This lets the same question be asked of the same task in five legal systems: how much of the answer is recoverable without a model, and what is left once it is removed?

Our contributions:
\begin{enumerate}[nosep]
    \item \textbf{A cross-jurisdictional resource}: five tasks mapped to registry metadata in five countries (six with the Ukrainian companion), with the same task definition, a frozen answer space per cell, native-language prompts and every model prediction released (Sections~\ref{sec:data}--\ref{sec:models}).
    \item \textbf{A leakage audit} with two instruments: a no-model keyword scan per cell, and a masked control that removes the label names from the input and re-asks the same model on the same documents (Sections~\ref{sec:leakage}--\ref{sec:masking}).
    \item \textbf{A paired comparison of models} that replaces per-model confidence intervals with McNemar tests on shared documents and a Holm correction, so that a cell has a leader only if its best model beats every other (Section~\ref{sec:paired}).
    \item \textbf{A correction record.} Auditing the harness surfaced defects in our own earlier releases, including one found for this version that overturns several v2 findings. We report each with its effect on the numbers (Section~\ref{sec:changes}).
\end{enumerate}

The practical recommendation is short. A benchmark built from a court registry should report, for every cell, the majority baseline, a no-model scan, and masked-input scores for several models of different strength. The first two cost nothing to compute; the third needs more than one model because models differ widely in how much they rely on the label names (Section~\ref{sec:masking}).

\section{Related Work}
\label{sec:related}

\paragraph{Shortcuts and leakage.}
Benchmarks can be solved by cues unrelated to the ability they are meant to measure. Hypothesis-only baselines solve a large share of natural language inference without reading the premise~\citep{gururangan2018annotation,poliak2018hypothesis}; models trained on such data adopt syntactic heuristics that fail on controlled counter-examples~\citep{mccoy2019right}; \citet{geirhos2020shortcut} frame the general pattern as shortcut learning. For LLMs a second channel is contamination: public test data may have been seen in pretraining~\citep{sainz2023contamination}. Court decisions are public by design, so both concerns apply to registry benchmarks. In prior work we showed that legal multiple-choice exams can be answered from the options alone, and that filtering items against one model does not remove the leakage for the next~\citep{ovcharov2026optionleakage}. The present paper concerns a different mechanism: the label is not inferable from the answer options but printed in the document.

\paragraph{Legal benchmarks.}
English benchmarks test common-law reasoning in one language: LegalBench~\citep{guha2024legalbench}, LexGLUE~\citep{chalkidis2022lexglue}, CUAD~\citep{hendrycks2021cuad}. Single-language benchmarks exist for Portuguese~\citep{canaverde2025legalbenchpt}, Vietnamese~\citep{dong2025vlegalbench}, Arabic~\citep{hijazi2024arablegaleval} and Japanese~\citep{fujita2025legalrikai}. LEXTREME~\citep{niklaus2023lextreme} spans 24 languages but evaluates \emph{different} tasks per language; its only same-task sub-tasks (MultiEURLEX, MAPA) draw on EU legislation rather than national court decisions. SCALE~\citep{niklaus2024scale} and cross-lingual judgment prediction~\citep{niklaus2022crossx} use court decisions in several languages but from one jurisdiction, Switzerland. \citet{ioannou2025limits} evaluate LLMs in 15 languages on official EU translations. LEXam~\citep{niklaus2025lexam} uses Swiss law-school exams. MultiLegalPile~\citep{niklaus2024multilegal} is a pretraining corpus without tasks. Table~\ref{tab:related_gap} places Multi-Legal-Bench against these. None of the benchmarks we know of reports a no-model baseline per task.

\begin{table}[t]
\centering
\caption{Comparison with existing legal benchmarks. ``Cross-juris.'' = the same task evaluated in several national legal systems. ``Native'' = decisions from national registries, not EU translations.}
\label{tab:related_gap}
\small
\begin{tabular}{lccccc}
\toprule
\textbf{Benchmark} & \textbf{Langs} & \textbf{Court dec.} & \textbf{Native} & \textbf{Cross-juris.} & \textbf{LLM eval} \\
\midrule
LegalBench          & 1 (EN)  & \xmark & -- & \xmark & \cmark \\
LexGLUE             & 1 (EN)  & \cmark & \cmark & \xmark & \xmark \\
LEXTREME            & 24      & 1 of 11 & partial & \xmark & \xmark \\
LEXTREME-MEU/MAPA   & 24      & \xmark & \xmark & \cmark & \xmark \\
SCALE               & 5       & \cmark & \cmark & \xmark & partial \\
Ioannou et al.      & 15      & \xmark & \xmark & \xmark & \cmark \\
\midrule
\textbf{Multi-Legal-Bench} & \textbf{6} & \cmark & \cmark & \cmark & \cmark \\
\bottomrule
\end{tabular}
\end{table}

\paragraph{Under-represented languages.}
No prior benchmark covers Ukrainian, Polish, Czech or Lithuanian legal text. Ukrainian is 0.5\% of mC4, eighteen times less than Russian~\citep{ovcharov2025crosslingual}. UA-Legal-Bench~\citep{ovcharov2025ualegalbench} is the Ukrainian-only companion of this benchmark; Multi-Legal-Bench extends its tasks to five further registries.

\section{Data}
\label{sec:data}

\subsection{Sources}
Each jurisdiction's decisions come from its official open registry: EDRSR for Ukraine~\citep{src_edrsr}, Judilibre for France~\citep{src_judilibre}, the Rechtspraak open data service for the Netherlands~\citep{src_rechtspraak}, the common-court and administrative-court portals for Poland~\citep{src_pl}, the Constitutional and Supreme Court databases for the Czech Republic~\citep{src_cz}, and LITEKO for Lithuania~\citep{src_liteko}. All are published under open government terms. They are part of the SecondLayer Legal Corpus, 165 million full-text decisions from 20 jurisdictions as of August 2026, of which Ukraine contributes 134 million (81\%).

\subsection{Jurisdiction selection}
\label{sec:selection}
We selected six of the 20 jurisdictions on three criteria: structured fields that support at least two of the five tasks; coverage of several language families (Slavic: Ukrainian, Polish, Czech; Romance: French; Germanic: Dutch; Baltic: Lithuanian); and enough full text to sample from (UA 133.6M, PL 2.83M, NL 0.92M, CZ 0.87M, FR 0.72M, LT 51K). Lithuania is small but is the only Baltic registry with the fields the tasks need. The US and UK were excluded for lack of structured labels, India for mixed-language text, and Latvia for lacking outcome and subject fields. Germany was excluded because its subject-area field is empty despite being present in the schema. We return to what this selection can and cannot support in Section~\ref{sec:discussion}: three of the six systems are Slavic-language, all six are civil-law, and no German-language legal system is included.

\subsection{Tasks and coverage}
\label{sec:tasks}

\begin{table}[t]
\centering
\caption{Task--jurisdiction coverage. Numbers are the answer classes the prompt offers. Per-cell sample statistics are in Table~\ref{tab:cellstats}.}
\label{tab:coverage}
\small
\begin{tabular}{lcccccc}
\toprule
\textbf{Task} & \textbf{UA} & \textbf{FR} & \textbf{NL} & \textbf{PL} & \textbf{CZ} & \textbf{LT} \\
\midrule
CTC & \cmark(4) & \cmark(3) & \cmark(6) & \cmark(6) & \cmark(3) & \cmark(4) \\
JFC & \cmark(4) & \cmark(6) & \cmark(2) & \cmark(3) & \cmark(3) & \xmark \\
COP & \cmark(6) & \cmark(7) & \xmark & \xmark & \xmark & \xmark \\
CCP & \cmark(19) & \cmark(10) & \cmark(11) & \xmark & \xmark & \cmark(10) \\
NE  & \cmark & \xmark & \xmark & \cmark & \cmark & \xmark \\
\bottomrule
\end{tabular}
\end{table}

The five tasks are inherited from UA-Legal-Bench~\citep{ovcharov2025ualegalbench}. \textbf{Court-type classification (CTC)} asks for the court's branch or level. \textbf{Judgment-form classification (JFC)} asks for the procedural form of the document (judgment, order, opinion). \textbf{Case-outcome prediction (COP)} gives only the facts section, with the ruling removed, and asks for the outcome; in this release it is available for France (the \emph{solution} field; 93\% of the sampled decisions are from the Cour de cassation and 7\% from courts of appeal, Table~\ref{tab:courts}). \textbf{Cause-category prediction (CCP)} asks for the subject area among macro-categories. \textbf{Norm extraction (NE)} asks for the cited provisions and is scored by set F1. Where metadata is absent we leave the cell empty rather than generate labels, so the matrix is sparse (Table~\ref{tab:coverage}).

The tasks are identical in definition, not in content. The answer spaces differ in size (CTC offers 3 classes in France and 6 in Poland) and in meaning, since each registry encodes its own court system. We therefore make Cohen's~$\kappa$~\citep{cohen1960coefficient}, which corrects for the number and balance of classes, the primary measure wherever cells are compared, and give accuracy beside it, and we fence the CCP and NE cells off from cross-jurisdictional claims: CCP cardinality ranges from 3 to 10 realised classes and NE ground truth comes from differently structured registry fields.

\paragraph{Sampling.}
Within each cell we sampled decisions dated 2020--2026, stratified by label where the registry allowed it, with 2{,}000--50{,}000 characters of text (Table~\ref{tab:cellstats}; the full label distribution is in Table~\ref{tab:labeldist} and the fields of a released record in Table~\ref{tab:schema}). Three exemplars per cell are drawn for the few-shot condition and held out of \emph{both} conditions. CCP documents whose registry category maps to none of the macro-categories the prompt offers cannot be answered and are excluded (FR 69, NL 1), rather than scored as errors. CZ~CTC realises only one class after filtering (all Constitutional Court); we report it for completeness and exclude it from every comparison.

\paragraph{Input.}
Models see the first 6{,}000 characters of the decision (3{,}000 of the facts section for COP), with the LITEKO portal header removed from Lithuanian documents (Section~\ref{sec:leakage}). The share of a median decision that this shows ranges from 36\% (FR~CCP) to 94\% (FR~COP; Table~\ref{tab:cellstats}). A control at a 24{,}000-character cap, which shows the median PL~CTC decision (22{,}345 characters) in full, moved PL~CTC by $-0.5$~pp on average (at most 2.6 for any of four models) and FR~CTC by $-0.05$~pp, so the cap does not drive the cross-jurisdictional ordering.

\paragraph{Court coding.}
To analyse institutional context (Section~\ref{sec:secondary}) every sampled decision is assigned the court that issued it and, through a coding scheme fixed before the analysis (Appendix Table~\ref{tab:scheme}), a level (first instance, appeal, apex, advisory, constitutional), a branch and a function (review of facts and law, or of law only). The court comes from the court code in the ECLI for NL and CZ, from the registry's own \emph{Teismas} field for LT, and from the court named in the decision's opening for FR and PL. Polish regional and Lithuanian county courts sit both at first instance and on appeal, so their level is read from the case number (Polish repertories such as \emph{Ca} and \emph{Ka}, Lithuanian prefixes such as \emph{e2A}). Checked by hand on samples of 50 decisions per country, the coding was correct in 50 of 50 for LT, and in 50 of 50 for FR and PL on fresh samples after one round of corrections (49 and 46 of 50 before them); 35 of 1{,}497 Polish decisions (2.3\%) remain without a level and are left out.

\section{Models and Evaluation}
\label{sec:models}

Table~\ref{tab:models} lists the models. The seven open-weight and second-tier hosted models of the first release ran through the Amazon Bedrock Converse API. Claude Sonnet~5 and Claude Haiku~4.5 were added for this version in answer to review; Bedrock had no Claude capacity available to us when the runs were made, so they ran through the Claude Code command-line client with every tool, extension and memory feature disabled, an empty working directory, the benchmark's own system prompt, and extended thinking turned off, so that they answer under the same conditions as the Bedrock models. GPT and Gemini models are not included; we discuss the consequence in Section~\ref{sec:limitations}.

\begin{table}[t]
\centering
\caption{Models. Fertility = tokens per whitespace word on Ukrainian legal text, measured with each model's own tokenizer (Section~\ref{sec:fertility}); Mistral Large 3, Ministral 3 and both Nemotron models share one tokenizer. MoE parameters are given as total/active. $^{*}$Not measured: the tokenizer is not public. $^{\dagger}$Not measured (gated repository); it shares the Llama~3 tokenizer of Llama~3.3.}
\label{tab:models}
\small
\resizebox{\linewidth}{!}{%
\begin{tabular}{llrrl}
\toprule
\textbf{Model} & \textbf{Provider} & \textbf{Parameters} & \textbf{Fert.\ (UA)} & \textbf{Access} \\
\midrule
Llama 4 Maverick~\citep{meta2025llama4}  & Meta     & 400B/17B & 2.30 & Bedrock \\
Llama 3.3 70B~\citep{grattafiori2024llama3} & Meta  & 70B      & 2.54 & Bedrock \\
Mistral Large 3~\citep{mistral2025large3} & Mistral & 675B/41B & 2.96 & Bedrock \\
Nemotron Super 3~\citep{nvidia2026nemotron3super} & NVIDIA & 120B/12B & 2.96 & Bedrock \\
Amazon Nova Pro~\citep{amazon2024nova}  & Amazon   & undisclosed & n/a$^{*}$ & Bedrock \\
Qwen3 235B~\citep{yang2025qwen3}        & Alibaba  & 235B/22B & 3.75 & Bedrock \\
Qwen3 32B~\citep{yang2025qwen3}         & Alibaba  & 32B      & 3.75 & Bedrock \\
Claude Sonnet 5~\citep{anthropic2026claude} & Anthropic & undisclosed & n/a$^{*}$ & Claude Code \\
Claude Haiku 4.5~\citep{anthropic2026claude} & Anthropic & undisclosed & n/a$^{*}$ & Claude Code \\
\midrule
\multicolumn{5}{l}{\emph{Small models, scaling analysis only (Appendix~\ref{app:scaling})}} \\
Ministral 3B / 8B~\citep{mistral2025large3} & Mistral & 3B / 8B & 2.96 & Bedrock \\
Llama 3.1 8B~\citep{grattafiori2024llama3}  & Meta    & 8B      & --$^{\dagger}$ & Bedrock \\
Nemotron Nano 12B v2 VL~\citep{nvidia2025nemotronnano2} & NVIDIA & 12B     & 2.96 & Bedrock \\
\bottomrule
\end{tabular}}
\end{table}

\paragraph{Protocol.}
Prompts are in the language of the jurisdiction, list the offered classes, and ask for a single label (Appendix~\ref{app:prompts}). Decoding is greedy (temperature~0) on Bedrock with a 512-token output limit; the Claude Code client does not expose temperature. Each cell is run zero-shot and with three exemplars. A response is mapped to the answer space by exact match after normalisation (lower case, diacritics folded, surrounding punctuation stripped), then by the longest offered label it contains; a response that matches nothing is an error, never a guess. The Bedrock runs used about 500M tokens; the 38 Claude runs added 95M. A Bedrock run took 1.5--4 minutes and a Claude run 3--11 minutes at four concurrent requests. Claude was run zero-shot only.

\paragraph{Statistics.}
All models answer the same documents, so model comparisons are paired. For each pair of models in a cell we test the discordant documents with McNemar's exact test~\citep{mcnemar1947note,dietterich1998approximate} and correct the $\binom{k}{2}$ tests of a cell with Holm's procedure~\citep{holm1979simple}. A cell has a \emph{leader} only if its most accurate model beats every other model at $\alpha = 0.05$ after correction. Because label spaces differ between cells, chance-corrected agreement (Cohen's~$\kappa$) is the primary measure for comparisons across cells; accuracy is reported beside it and is used within a cell, where all models face the same label space.

\section{What a Model-Free Scan Recovers}
\label{sec:leakage}

We compute two baselines that use no model (Table~\ref{tab:baselines}). \textbf{Majority} predicts the most frequent label. \textbf{Scan} reads the first 1{,}500 characters of exactly the text the model is shown and returns the longest offered label whose name appears there, falling back to the majority label. It has no parameters, no prompt and no training.

\begin{table}[t]
\centering
\caption{Model-free baselines against the models, zero-shot, on the documents all models answered. $\kappa$ is the primary measure because the cells differ in the number and balance of classes. \emph{Scan}: accuracy and $\kappa$ of the keyword scan; \emph{Bedrock}: range over the seven Bedrock models; \emph{Claude}: $\kappa$ of Sonnet~5 / Haiku~4.5. \textbf{Bold}: the scan solves most of the cell. $\ddagger$~degenerate, one realised class.}
\label{tab:baselines}
\small
\setlength{\tabcolsep}{4pt}
\begin{tabular}{lrrrrrrl}
\toprule
& & & \multicolumn{2}{c}{\textbf{Scan}} & \multicolumn{2}{c}{\textbf{Bedrock}} & \\
\cmidrule(lr){4-5} \cmidrule(lr){6-7}
\textbf{Cell} & \textbf{\#cls} & \textbf{Maj} & \textbf{Acc} & \textbf{$\kappa$} & \textbf{$\kappa$} & \textbf{Acc} & \textbf{Claude $\kappa$} \\
\midrule
FR CTC & 3 & 33 & 84 & \textbf{0.76} & 0.99--1.00 & 99--100 & 1.00 / 0.99 \\
FR JFC & 6 & 21 & 45 & 0.32 & 0.43--0.72 & 54--78 & 0.69 / 0.54 \\
FR COP & 7 & 32 & 32 & 0.00 & 0.15--0.25 & 30--41 & 0.31 / 0.13 \\
FR CCP & 10 & 34 & 11 & -0.06 & 0.46--0.56 & 56--67 & 0.55 / 0.46 \\
NL CTC & 6 & 17 & 17 & 0.00 & 0.78--0.96 & 82--97 & 0.94 / 0.90 \\
NL JFC & 2 & 50 & 96 & \textbf{0.93} & 0.99--1.00 & 99--100 & 1.00 / 1.00 \\
NL CCP & 3 & 47 & 47 & 0.00 & 0.21--0.55 & 29--65 & 0.37 / 0.46 \\
PL CTC & 3 & 33 & 33 & 0.00 & 0.36--0.43 & 52--60 & 0.35 / 0.33 \\
PL JFC & 3 & 33 & 66 & \textbf{0.50} & 0.51--0.61 & 67--74 & 0.55 / 0.57 \\
CZ CTC$^{\ddagger}$ & 1 & 100 & 100 & -- & -- & 100 & -- \\
CZ JFC & 3 & 50 & 56 & 0.13 & 0.41--0.86 & 70--93 & 0.91 / 0.36 \\
LT CTC & 4 & 25 & 49 & 0.32 & 0.64--0.98 & 73--99 & 0.93 / 0.81 \\
LT CCP & 10 & 59 & 59 & 0.00 & 0.36--0.47 & 53--62 & 0.46 / 0.38 \\
 
\bottomrule
\end{tabular}
\end{table}

The scan reaches 96\% on NL~JFC against a 50\% majority ($\kappa = .93$), 84\% on FR~CTC against 33\% ($\kappa = .76$), and 66\% on PL~JFC against 33\% ($\kappa = .50$); on PL~JFC the seven Bedrock models add only 1 to 7 points to it. The alignment with apparent difficulty is close but not exact. Of the three cells the scan largely solves, FR~CTC and NL~JFC are near ceiling for every model and PL~JFC is where the models barely exceed the scan; the cells that are hard for every model (PL~CTC, FR~COP and the CCP cells) are ones where the scan sits at or below the majority floor. NL~CTC is the informative exception: the scan recovers nothing ($\kappa = 0$), yet the models reach $\kappa = .78$--$.96$, so a cell can be both free of name leakage and easy. Three cells fall between: on FR~JFC, CZ~JFC and LT~CTC the scan recovers something (45, 56 and 49\%) and the models clearly go beyond it.

The leakage has two sources. Most of it is intrinsic to the documents: a decision states the name of the court that issued it and the form it takes, because that is what it is. One source was an artefact of collection and is removed in this release: every Lithuanian document arrived with a LITEKO portal header that printed the case type and the cause categories outright. Stripping the header cut the leak from all 1{,}000 LT documents to 58 and dropped the LT~CTC scan from 75\% to 49\%, and LT~CTC and LT~CCP were re-run on the stripped text. The intrinsic leakage cannot be removed without rewriting the documents, which is why we measure what happens when we try (Section~\ref{sec:masking}).

Two cells are uninformative for a reason other than leakage. LT~CCP has a 59\% majority class and five of the seven Bedrock models score below it, so its 53--62\% accuracy says little. CZ~CTC realises a single class.

\section{Masking the Label Names}
\label{sec:masking}

A scan shows that the answer is \emph{available}; it does not show that a model \emph{uses} it. To test that, we re-ask the same model on the same documents with the label names removed. For each of the six cells where the scan exceeds the majority baseline, a lexicon of the offered label names and their inflections (e.g.\ Polish \emph{wyrok}, \emph{wyroku}, \emph{wyrokiem}; Czech \emph{n\'alez}, \emph{n\'alezu}; French \emph{cour d'appel}, \emph{appel}) is replaced by the token \texttt{[MASK]} after truncation, so the model sees the same span of text. In NL~JFC the court code inside the ECLI identifier is masked as well, since it names the issuing body. Masking removes 6 to 10 occurrences per document on average.

The masking is verified before it is used: after masking, the scan falls to the majority floor or below in all six cells (NL~JFC $96.5 \to 51.2$\%, FR~CTC $83.9 \to 33.2$\%, PL~JFC $66.5 \to 33.2$\%, LT~CTC $48.8 \to 25.1$\%, FR~JFC $45.0 \to 16.9$\%, CZ~JFC $56.3 \to 45.8$\%). What masking does \emph{not} remove is deliberate: structural cues that correlate with the label without naming it, such as a Lithuanian case-number prefix or, in the Dutch metadata, the name of the Procurator General's office that issues every \emph{conclusie}. Accuracy that survives masking is therefore an upper bound on what a cell measures beyond name lookup, not a measure of legal reasoning.

Table~\ref{tab:masked} summarises the result for all nine models and Figure~\ref{fig:masked} shows it model by model (per-model values in Appendix Table~\ref{tab:masked_full}). Masking costs most models a great deal. After a Holm correction across the 54 model--cell pairs, accuracy drops significantly in 39, and eight of the nine models lose significantly in at least four of the six cells, on average by 7 to 21 points (Qwen3~32B: from 99.8\% to 52.3\% on NL~JFC). One model is the exception: Claude Sonnet~5 stays within one point of its unmasked accuracy in five cells, and its one significant drop is 4.2~points on CZ~JFC. The label names therefore carry much of what most models achieve in these cells, and a model that does not need them exists.

Three consequences follow. First, the scan and the models answer different questions. On FR~CTC the scan solves 84\% of the cell, yet the median model loses only 1.1~points when the names are masked; on NL~JFC the scan and most models alike depend on the names. Whether a cell measures lookup is a property of the model--cell pair, not of the cell. Second, agreement between a model and the scan does not show that the model \emph{is} a scan. On PL~JFC the models barely exceed the scan's 66.5\%, which in v2 we read as lookup; once masked, the scan falls to 33\% while the nine models keep 57--71\%, so the ceiling on that cell is the difficulty of telling a judgment from a judgment with reasons. Third, leakage compresses the models towards the ceiling. The spread between the best and the worst model widens in every masked cell, from 0.6 to 47.6 points on NL~JFC, from 25.7 to 45.4 on LT~CTC and from 23.3 to 43.1 on FR~JFC. Section~\ref{sec:paired} returns to what this means for comparing models.

\begin{figure}[t]
\centering
\input{figures/fig_masked.tex}
\caption{Masked control. For each model and leaky cell, zero-shot accuracy with the label names in the input (grey) and masked (circle); a red circle marks a significant drop (McNemar, Holm across the 54 model--cell pairs). Only Claude Sonnet~5 moves little across the six cells (its one significant drop is 4.2 points on CZ~JFC).}
\label{fig:masked}
\end{figure}

\begin{table}[t]
\centering
\caption{Masked control, zero-shot, nine models. \emph{Scan}: keyword scan before $\to$ after masking. $\Delta$: masked minus plain accuracy, same model and documents. \emph{Sig.}: models whose accuracy drops significantly (McNemar, Holm across all 54 model--cell pairs). Per-model values in Appendix Table~\ref{tab:masked_full}.}
\label{tab:masked}
\small
\setlength{\tabcolsep}{4pt}
\begin{tabular}{lrrrrrr}
\toprule
\textbf{Cell} & \textbf{Scan} & \textbf{Median $\Delta$} & \textbf{Range of $\Delta$} & \textbf{Sig.} & \textbf{Sonnet~5} & \textbf{Haiku~4.5} \\
\midrule
CZ JFC & 56.3 $\to$ 45.8 & $-$15.5 & $-$26.0 to $-$2.4 & 8/9 & $-$4.2 & $-$8.0 \\
FR CTC & 83.9 $\to$ 33.2 & $-$1.1 & $-$6.7 to +0.0 & 4/9 & +0.0 & $-$2.0 \\
FR JFC & 45.0 $\to$ 16.9 & $-$13.9 & $-$30.1 to $-$0.8 & 7/9 & $-$0.8 & $-$14.5 \\
LT CTC & 48.8 $\to$ 25.1 & $-$14.8 & $-$27.2 to +0.9 & 8/9 & +0.9 & $-$18.7 \\
NL JFC & 96.5 $\to$ 51.2 & $-$15.6 & $-$47.5 to $-$0.1 & 6/9 & $-$0.1 & $-$33.1 \\
PL JFC & 66.5 $\to$ 33.2 & $-$8.3 & $-$11.6 to +1.2 & 6/9 & $-$0.8 & $-$5.6 \\
 
\bottomrule
\end{tabular}
\end{table}

Haiku's losses on LT~CTC and CZ~JFC are partly a formatting effect. Without the label name it more often answers with a paragraph instead of one word (LT~CTC: 89 multi-line answers unmasked, 226 masked; CZ~JFC: 370 and 447 of 666), and a response is read from its first line only, as for every model (Appendix Table~\ref{tab:format}). The drops on FR~JFC and NL~JFC, where Haiku's answers stay single-word (2 and 4, 0 and 16 multi-line), are not formatting effects. Two Bedrock models show the same shift in one cell each: Llama~4 Maverick on FR~CTC (65 to 365 multi-line answers, with a loss of 6.0~points that is therefore partly one of format) and Nova Pro on NL~JFC (2 to 80, with no loss); elsewhere the Bedrock models' answers stay on one line.

\section{Comparing Models}
\label{sec:paired}

\begin{table}[t]
\centering
\caption{Paired comparison, zero-shot. Top-two models by accuracy, their gap (pp), the discordant documents (top right and second wrong / the reverse), McNemar exact $p$ for the pair, whether the top model beats \emph{every} other model after Holm correction, and how many model pairs in the cell differ significantly after correction.}
\label{tab:paired}
\small
\setlength{\tabcolsep}{4pt}
\begin{tabular}{lrllrrrcr}
\toprule
\textbf{Cell} & \textbf{$n$} & \textbf{Top} & \textbf{Second} & \textbf{Gap} & \textbf{Disc.} & \textbf{$p$} & \textbf{Leader} & \textbf{Sig.\ pairs} \\
\midrule
FR CTC & 996 & Qwen3 235B & Sonnet 5 & 0.0 & 1/1 & 1.00 & \xmark & 0/36 \\
FR JFC & 787 & Qwen3 235B & Sonnet 5 & 2.3 & 24/6 & 0.001 & \cmark & 32/36 \\
FR COP & 810 & Sonnet 5 & Qwen3 235B & 3.2 & 173/147 & 0.16 & \xmark & 18/36 \\
FR CCP & 928 & Nemotron S3 & Nova Pro & 1.5 & 50/36 & 0.16 & \xmark & 21/36 \\
NL CTC & 993 & Mistral L3 & Qwen3 32B & 0.4 & 23/19 & 0.64 & \xmark & 22/36 \\
NL JFC & 998 & Maverick & Llama 3.3 & 0.0 & 0/0 & 1.00 & \xmark & 0/36 \\
NL CCP & 996 & Qwen3 32B & Nemotron S3 & 1.8 & 71/53 & 0.13 & \xmark & 31/36 \\
PL CTC & 495 & Maverick & Nemotron S3 & 0.4 & 5/3 & 0.73 & \xmark & 22/36 \\
PL JFC & 996 & Nemotron S3 & Maverick & 1.4 & 122/108 & 0.39 & \xmark & 4/36 \\
CZ JFC & 666 & Sonnet 5 & Llama 3.3 & 2.4 & 36/20 & 0.04 & \xmark & 28/36 \\
LT CTC & 997 & Mistral L3 & Nova Pro & 3.8 & 39/1 & $<$0.001 & \cmark & 32/36 \\
LT CCP & 961 & Mistral L3 & Maverick & 0.5 & 33/28 & 0.61 & \xmark & 24/36 \\
 
\bottomrule
\end{tabular}
\end{table}

Table~\ref{tab:paired} replaces the per-model confidence intervals of the first release with paired tests. The accuracy and $\kappa$ of every model in every cell are in Appendix Tables~\ref{tab:full_acc} and~\ref{tab:full_kappa}, and Figure~\ref{fig:model_rankings} shows the zero-shot $\kappa$ of all nine models. Two results stand out. First, models do differ: 234 of the 432 model pairs differ significantly within their cell after correction. Second, the differences do not add up to a leader. Only two of the twelve informative cells have a model that beats every other: Qwen3~235B on FR~JFC (2.3~pp ahead of Claude Sonnet~5, 24 discordant documents against 6) and Mistral Large~3 on LT~CTC (3.8~pp ahead of Nova Pro, 39 against 1). In the 3-shot condition there are again two, and they are different models on different cells: Nemotron Super~3 on PL~CTC and Qwen3~32B on PL~JFC (Appendix Table~\ref{tab:paired_fs}). No model leads in more than one cell under either condition. Adding Claude changes neither leader. Sonnet~5 is the most accurate model on FR~COP (44.2\%) and CZ~JFC (95.5\%), but on FR~COP it does not differ significantly from the next model, and on CZ~JFC it does not beat Llama~3.3~70B after correction ($p = 0.04$ before). Haiku~4.5 falls inside or below the Bedrock range in every cell. The 3-shot comparison covers the seven Bedrock models only.

The masked control qualifies this parity. On the six cells where label names leak, the same paired test run on the masked inputs puts Claude Sonnet~5 first in five, and in the two cells with a leader, FR~JFC and LT~CTC, it is Sonnet~5 that beats every other model (by 8.8 and 7.1~points over the next model), not the unmasked leaders Qwen3~235B and Mistral Large~3, which lose 11.8 and 10.0~points without the names. The leakage lifts weaker models towards the ceiling and so hides differences that exist. Parity on the unmasked cells is therefore partly an artefact of the cells, and the clean cells (NL~CTC, PL~CTC, FR~COP and the CCP cells), where no leader emerges either, are the better evidence for it.

The paired analysis changes one conclusion of the first release. That version reported three cells with a significant leader, among them Nemotron Super~3 on CZ~JFC at $z = 7.5$. The CZ~JFC result was produced by the scoring defect described in Section~\ref{sec:changes}: five of seven models were answering in Czech spelling and were scored as wrong. With the defect removed the most accurate Bedrock model on CZ~JFC is Llama~3.3~70B (93.1\%), which does not beat Mistral Large~3 after correction; with Claude added, Sonnet~5 is ahead of both (95.5\%) and does not beat Llama~3.3 after correction either.

We draw a narrower conclusion than the title of the first release did. On these cells, with these models, the variation between cells (the best model's $\kappa$ ranges from .31 on FR~COP to 1.00 on NL~JFC) is an order of magnitude larger than the variation between the best models within a cell, and which model is best depends on the cell. That is a statement about this model set and these tasks, measured on inputs that contain the answer in six of twelve cells. It is not evidence that no model could dominate: the masked comparison shows that leakage can hide a leader, although in the cells that never leaked none emerges.

\section{Secondary Analyses}
\label{sec:secondary}

The analyses in this section are exploratory. Each compares a small number of cells whose tasks, label spaces and courts differ at the same time, so none of them isolates a cause.

\paragraph{Few-shot effects.}
Three exemplars change accuracy by $-13.5$ to $+41.8$~pp across model--cell pairs (Figure~\ref{fig:fewshot_delta}), and the direction is not predictable from the task, the language or the headroom: the correlation between a cell's mean gain and the room its zero-shot score leaves is $r = 0.18$ over 12 cells, not significant. v2 reported $r = 0.59$ and a mean gain of $+35.8$~pp on CZ~JFC, both driven by the scoring defect; on correctly scored predictions CZ~JFC loses 1.9~pp on average.

\begin{figure}[t]
\centering
\input{figures/fig2_fewshot_delta.tex}
\caption{Few-shot effect per cell: 3-shot minus zero-shot accuracy (pp), one grey dot per Bedrock model, red diamond the mean over the seven (CZ~CTC, one class, omitted). No task gains uniformly; the sign varies within cells.}
\label{fig:fewshot_delta}
\end{figure}

\paragraph{Cross-lingual exemplars.}
Replacing native exemplars with Ukrainian ones costs $-2.0$~pp on FR~CTC and $-13.8$~pp on PL~CTC (Table~\ref{tab:xling}), the opposite of what language proximity would predict. The cells that transfer at or near zero have answer spaces that map onto Ukraine's (LT's four court types onto Ukraine's four, FR's three onto a coarsening of them); PL~CTC's six court types do not. This is consistent with label-set alignment mattering more than language family, but five target cells and two tasks cannot separate the two. We also note, following review, that linguistic proximity and legal proximity are different things: Polish and Ukrainian are close languages whose court systems are organised differently, and nothing in this design would detect legal proximity if it were the operative factor.

\begin{table}[t]
\centering
\caption{Ukrainian vs native exemplars, 3-shot, mean over the seven Bedrock models, strictly paired by document.}
\label{tab:xling}
\small
\begin{tabular}{ll R{1.1cm} R{1.1cm} R{1.1cm} R{1.1cm} R{1.1cm} R{1.1cm}}
\toprule
& & \multicolumn{3}{c}{\textbf{CTC}} & \multicolumn{3}{c}{\textbf{JFC}} \\
\cmidrule(lr){3-5} \cmidrule(lr){6-8}
\textbf{Target} & \textbf{Family} & UA & Native & $\Delta$ & UA & Native & $\Delta$ \\
\midrule
PL & Slavic    & 40.1 & 53.9 & $-$13.8 & 66.5 & 68.5 & $-$2.0 \\
CZ & Slavic    & 100  & 100  & 0.0     & 67.8 & 77.0 & $-$9.2 \\
LT & Baltic    & 92.6 & 92.8 & $-$0.2  & --   & --   & -- \\
FR & Romance   & 97.8 & 99.8 & $-$2.0  & 63.4 & 72.5 & $-$9.1 \\
NL & Germanic  & 91.4 & 97.4 & $-$6.0  & 99.9 & 98.7 & $+$1.1 \\
\bottomrule
\end{tabular}
\end{table}

\paragraph{Tokenizer fertility.}
\label{sec:fertility}
Tokenizer fertility is the average number of subword tokens a tokenizer produces per whitespace-delimited word~\citep{rust2021tokenizer}; higher fertility means a language is split into more pieces and costs more to process. We measured it for nine models on 100 decisions per language, each with the model's own tokenizer and with the Lithuanian portal header removed (Appendix Figure~\ref{fig:fertility_heatmap}). The nine share only four tokenizers: Mistral Large 3, the Ministral 3 models and both Nemotron models use the same one, and the two Qwen3 models another. French is the cheapest language for every tokenizer (1.70--1.99) and Lithuanian the most expensive (2.97--3.62); in between, the order depends on the tokenizer (Polish costs more than Ukrainian for Llama~3, Czech more than Polish for Qwen3). Across 71 model--cell pairs, fertility is not associated with zero-shot accuracy ($r = -0.07$, $p = 0.54$), but with four distinct tokenizers these pairs are far from independent, so the test has little power. Fertility predicts cost, since tokens are billed; we find no sign that it predicts quality across languages.

\paragraph{Court level.}
Whether identically named tasks hide different institutional tasks can be tested directly with the court coding. Two findings stand out. First, in two cells the label is largely a function of the court. Every NL~JFC \emph{conclusie} comes from the Procurator General and every \emph{uitspraak} from a court, and in FR~JFC the classes \emph{qpc}, \emph{avis} and \emph{saisie} occur only at the Cour de cassation. The FR~JFC class \emph{autre} is the registry's residual category: it labels 52 of the 54 first-instance decisions in the cell and 112 decisions of the Cour de cassation that are, in substance, \emph{arr\^ets}. Every model scores 0--2.4\% on these 165 documents (21\% of the cell), answering \emph{arr\^et} for 109 to 122 of them; without them FR~JFC accuracy is 68.6--98.1\%. Much of what made FR~JFC look hard is a registry category no model can be expected to reproduce, the kind of label-validity problem that expert annotation would catch.

Second, within the same gold label, accuracy does depend on the level of the court (Table~\ref{tab:courtlevel}). The effects run in both directions and are not a simple ``higher courts are harder'' pattern: Dutch appellate decisions are 12.7~points harder to assign a subject area than first-instance ones, while Lithuanian appellate and apex decisions are easier to assign a case type (by 5.9 and 2.3~points). On FR~COP the 59 decisions of courts of appeal are 13.9~points harder than those of the Cour de cassation, whose outcome categories (\emph{cassation}, \emph{rejet}) the task was built around. Three of the nine contrasts with a regression estimate survive a Holm correction across them. Because the cells mix levels in different proportions (FR~CCP is 89\% courts of appeal, NL~CCP 70\% first instance), a comparison of the same task across jurisdictions is also a comparison across institutional mixes; this is a further reason the cross-jurisdictional analyses are exploratory.

\begin{table}[t]
\centering
\caption{Accuracy by court level, zero-shot, within gold-label strata (so that a different label mix at each level cannot pass for an effect), averaged over the nine models; 95\% interval from 1{,}000 resamples of documents within strata. $n$: documents at the level / at the reference level in the shared strata. $p$: court-level term in a logistic regression of correctness on level, model and gold label with document-clustered errors; Holm: adjusted across the nine contrasts with a regression estimate (the FR~JFC regressions do not converge because the level nearly determines the label). NL~JFC admits no within-label contrast at all.}
\label{tab:courtlevel}
\small
\setlength{\tabcolsep}{3pt}
\resizebox{\linewidth}{!}{%
\begin{tabular}{llrrrrrr}
\toprule
\textbf{Cell} & \textbf{Contrast} & \textbf{$n$} & \textbf{$\Delta$ (pp)} & \textbf{95\% interval} & \textbf{Models} & \textbf{$p$} & \textbf{Holm} \\
\midrule
FR JFC & first instance vs apex & 52 / 112 & +0.4 & [$-$0.2, +1.3] & +0.0 to +2.1 & -- & -- \\
FR JFC & appeal vs apex & 117 / 161 & $-$0.6 & [$-$1.1, $-$0.1] & $-$2.5 to +0.0 & -- & -- \\
FR CCP & apex vs appeal & 63 / 842 & +9.1 & [+0.4, +17.0] & +4.2 to +16.7 & 0.07 & 0.21 \\
FR COP & appeal vs apex & 59 / 209 & $-$13.9 & [$-$21.4, $-$5.0] & $-$26.2 to $-$3.3 & 0.010 & 0.06 \\
NL CCP & appeal vs first instance & 145 / 694 & $-$12.7 & [$-$18.3, $-$6.6] & $-$17.4 to $-$7.9 & $<$0.001 & $<$0.001 \\
NL CCP & apex vs first instance & 147 / 694 & +1.4 & [$-$5.5, +9.0] & $-$12.3 to +8.0 & 0.56 & 0.77 \\
PL JFC & appeal vs first instance & 368 / 604 & $-$2.2 & [$-$4.2, $-$0.1] & $-$10.4 to +0.5 & 0.05 & 0.18 \\
LT CTC & appeal vs first instance & 203 / 494 & +5.9 & [+4.4, +7.5] & $-$5.1 to +16.4 & $<$0.001 & $<$0.001 \\
LT CTC & apex vs first instance & 167 / 620 & +2.3 & [+1.0, +3.7] & $-$3.3 to +7.1 & 0.006 & 0.04 \\
LT CCP & appeal vs first instance & 262 / 647 & +6.1 & [+0.5, +11.4] & +0.3 to +10.3 & 0.03 & 0.13 \\
LT CCP & apex vs first instance & 33 / 642 & $-$6.1 & [$-$14.7, +2.0] & $-$12.3 to $-$0.6 & 0.38 & 0.77 \\
 
\bottomrule
\end{tabular}}
\end{table}

\paragraph{Memorisation.}
The decisions are public, so a model might answer from pretraining memory rather than from the text. We cannot inspect pretraining data, but a decision published after a model's training cutoff cannot have been memorised. Within each cell and gold label, so that a shift in the label mix across years cannot pass for an effect, we compare zero-shot accuracy on decisions dated up to 2023 (\temporalnpre{} documents) with decisions from 2025 on (\temporalnpost{}), pooling the per-stratum differences and resampling documents for a 95\% interval (Table~\ref{tab:temporal}). Every model is 0.3 to 3.0 points \emph{less} accurate on the recent decisions. Memorisation predicts a larger gap for models that could have seen the older decisions but not the recent ones than for models whose training covers both. The two models with published cutoffs that make every 2025 decision unseen, Llama~3.3~70B (December 2023) and Llama~4 Maverick (August 2024), lose 2.4 and 2.1 points, inside the range of the other seven models (0.3 to 3.0) and with overlapping intervals; Sonnet~5, one of the most recent models, loses the most (3.0). The data give no sign of that pattern, and the small common decline is more plausibly a change in the decisions themselves. This bounds memorisation rather than excluding it: a model could have memorised older decisions and still be weaker on newer ones for other reasons.

\begin{table}[t]
\centering
\caption{Accuracy on decisions from 2025 on minus accuracy on decisions up to 2023, zero-shot, pooled within cell $\times$ gold-label strata; 95\% interval from 1{,}000 resamples of documents within strata.}
\label{tab:temporal}
\small
\begin{tabular}{lrr}
\toprule
\textbf{Model} & \textbf{$\Delta$ (pp)} & \textbf{95\% interval} \\
\midrule
Maverick & $-$2.1 & [$-$4.2, $-$0.2] \\
Llama 3.3 & $-$2.4 & [$-$4.3, $-$0.4] \\
Mistral L3 & $-$2.4 & [$-$4.1, $-$0.7] \\
Nemotron S3 & $-$0.3 & [$-$2.1, +1.7] \\
Nova Pro & $-$1.4 & [$-$3.4, +0.4] \\
Qwen3 235B & $-$1.9 & [$-$3.7, $-$0.0] \\
Qwen3 32B & $-$1.4 & [$-$3.2, +0.4] \\
Sonnet 5 & $-$3.0 & [$-$4.9, $-$0.9] \\
Haiku 4.5 & $-$2.7 & [$-$4.5, $-$0.8] \\
 
\bottomrule
\end{tabular}
\end{table}

\paragraph{Scale.}
Four models of 3--12B parameters trail the mean of the seven Bedrock models by 3 to 21~pp in most cells (Appendix Table~\ref{tab:scaling}), most on CZ~JFC ($-21$) and LT~CTC ($-21$) and least on FR~COP ($-3$), where every model is weak. NL~CCP is the exception ($+5$), driven by Nemotron Nano~12B.

\section{Discussion}
\label{sec:discussion}

\paragraph{What registry cells measure.}
A registry field is a label, but it is also a fact about the document that the document usually states. The scan sorts the cells into those where the answer is printed (FR~CTC, NL~JFC, PL~JFC), those where it is printed but insufficient (FR~JFC, CZ~JFC, LT~CTC) and those where it is absent (NL~CTC, PL~CTC, FR~COP and the CCP cells). The last group holds the cells where models are weakest and differ most, with NL~CTC as the exception that models solve without the name. The masked control complicates the first two groups. For eight of nine models the printed label carries much of the score, up to half of it on NL~JFC, while for Sonnet~5 it carries almost nothing, so a high score in a leaky cell is ambiguous until a masked score is reported beside it. Even in the clean cells a model may rely on cues we have not identified. The general lesson is not specific to our data: any benchmark that maps a task onto a registry field should state, cell by cell, how much of the field the input already contains.

\paragraph{Linguistic and legal proximity.}
Several of our comparisons group jurisdictions by language family. A lawyer reading them would group differently: by legal family, by the institutional role of the court, and by which foreign courts a system actually cites. Czech and Polish are close languages, but a Czech Constitutional Court \emph{n\'alez} and a Polish district court \emph{wyrok} are different kinds of act. The same task name can therefore hide different legal tasks. The clearest case is outcome prediction on French decisions: the Cour de cassation reviews the application of law, not the facts, so predicting \emph{cassation} or \emph{rejet} from a facts section is a harder and in part different task than predicting the outcome of a first-instance decision. The composition of each cell by court (Table~\ref{tab:courts}) shows how much institutional structure hides inside a task label. Half of NL~JFC consists of opinions of the Procurator General at the Hoge Raad, and every \emph{conclusie} in the cell comes from that office, so the task is in large part ``which institution wrote this''; that is how the models that survive masking there still answer, from the office's name in the metadata. CZ~JFC is drawn entirely from the Constitutional Court, so it separates a judgment on the merits (\emph{n\'alez}) from a procedural resolution (\emph{usnesen\'i}) within one court. FR~CCP is 89\% courts of appeal and FR~COP 93\% Cour de cassation. Coding the level of each court (Section~\ref{sec:secondary}) confirms that this structure matters: within the same label, accuracy moves by up to 14 points with the level of the court, in both directions, and in FR~JFC a residual registry category attached to particular courts accounts for most of the models' errors. What the current samples cannot provide is a comparison of jurisdictions at matched levels, since the cells mix levels in different proportions; a sample stratified by level is the next step for the COP and CCP cells.

\paragraph{Which jurisdictions would test the hypotheses.}
The selection limits what the cross-lingual results can show. Three of six systems are Slavic-language, and the German-language legal systems, which several of the included jurisdictions treat as a point of reference, are absent. A German-language triad (Germany, Austria, Switzerland) would separate the effect of a shared language from that of distinct but interacting legal systems, and a same-system comparison in two languages (the Swiss Federal Supreme Court publishes in German, French and Italian) would do the converse. Our corpus now holds 1.2 million Swiss federal and cantonal decisions, and we intend these as the next extension.

\section{Limitations}
\label{sec:limitations}

\textbf{Model coverage.} The models include Claude Sonnet~5 and Haiku~4.5 but no GPT or Gemini model, so a model that dominates these cells cannot be ruled out; the paired analysis describes this model set only. \textbf{Masked control.} Masking removes label names but not structural cues, and so bounds rather than measures the non-lookup content of a cell; it was run zero-shot only. \textbf{Contamination.} The decisions are public and some were plausibly in pretraining data. The temporal comparison (Table~\ref{tab:temporal}) finds no sign that models are more accurate on decisions they could have seen, but it bounds memorisation rather than excluding it, and only two models have published cutoffs early enough that every 2025 decision is unseen. \textbf{Answer format.} Responses are read from their first line, the same rule for every model; Haiku~4.5 often answers in several lines (up to 370 of 666 on CZ~JFC), which lowers its scores and part of its masked losses for reasons of format rather than content. \textbf{Court coding.} The court is read from the opening of the decision for FR and PL; checked on samples of 50 it was correct in every case after corrections, but not every decision was checked, and 2.3\% of Polish decisions have no level. \textbf{Labels.} Gold labels are registry metadata, not expert annotation, and may contain errors. \textbf{Jurisdictions.} All six systems are European and civil-law. \textbf{Decoding.} The Claude Code client does not expose temperature, so Claude runs are not strictly deterministic. \textbf{Input length.} Models see a 6{,}000-character prefix; the truncation control covers two cells and four models.

\section{Changes from v2}
\label{sec:changes}

This version was prepared after peer review of a workshop version and after a further audit of the scoring code. Several results of v2 change.

\begin{enumerate}[nosep]
\item \textbf{Diacritics in scoring.} The v2 scorer compared responses with the ASCII answer space (\texttt{usneseni}, \texttt{nalez}, \texttt{arret}) without folding diacritics, so a model that answered \emph{usnesen\'i} was scored wrong. Five of seven models answer Czech judgment form in Czech spelling. CZ~JFC zero-shot, reported as 0.3--74.8\%, is 69.8--93.1\%; FR~JFC moves by up to 9.0~pp for two models. The same defect was present in the workshop version.
\item \textbf{Retracted:} a significant leader on CZ~JFC (Nemotron Super~3, $z = 7.5$); the claim that Czech judgment-form labels are opaque to models without examples; the mean few-shot gain of $+35.8$~pp on CZ~JFC (it is $-1.9$); the benefit of Ukrainian exemplars over zero-shot on CZ~JFC (they lower accuracy); and the correlation of few-shot gain with headroom ($r = 0.59$; it is $0.18$).
\item \textbf{Statistics.} Per-model binomial intervals are replaced by paired McNemar tests with Holm correction; Cohen's~$\kappa$ is the primary measure across cells.
\item \textbf{New experiments.} Claude Sonnet~5 and Haiku~4.5 on all cells, and the masked control for all nine models (Section~\ref{sec:masking}), which also changes the reading of the model comparison (Section~\ref{sec:paired}).
\item \textbf{Framing.} Label leakage, one of five findings in v2, is the subject of this version. The claim that no model dominates is restated as a property of this model set.
\item \textbf{Reporting.} Per-cell data statistics (length, full label distribution, court composition), the record schema, full per-model results including norm extraction, model sizes, all prompts with a complete 3-shot example, a definition of tokenizer fertility, a temporal check for memorisation, a court-level analysis with its coding scheme, tokenizer fertility re-measured with each model's own tokenizer (v2 used stand-in tokenizers for four models without saying so), and references for the models and data sources are added.
\end{enumerate}

\section{Conclusion}
\label{sec:conclusion}

Registry metadata makes legal benchmarks cheap to build, and it makes some of their cells cheap to solve. In Multi-Legal-Bench, the cells that looked easiest are the ones where a keyword scan without a model does most of the work, and masking the label names costs eight of nine models up to 48 points while one, Claude Sonnet~5, loses at most 4. With the names in place no model leads in more than one cell; with them masked, one model leads in both masked cells that have a leader. Leakage does not only inflate scores, it also hides the differences between models. We recommend three numbers for every cell of a registry-derived benchmark: the majority baseline, a model-free scan, and masked-input scores for more than one model, since a single model's masked score can hide what the others depend on.

\paragraph{Data availability.}
Data, prompts, predictions and scoring code: \url{https://huggingface.co/datasets/overthelex/multi-legal-bench}.

\section*{Acknowledgments}
We thank the three anonymous reviewers of the NLLP 2026 workshop, whose comments shaped this version: the paired analysis, the masked control and the court-level analysis all respond to their reviews.

\bibliographystyle{plainnat}
\bibliography{references}

\begin{thebibliography}{40}
\providecommand{\natexlab}[1]{#1}
\providecommand{\url}[1]{\texttt{#1}}
\expandafter\ifx\csname urlstyle\endcsname\relax
  \providecommand{\doi}[1]{doi: #1}\else
  \providecommand{\doi}{doi: \begingroup \urlstyle{rm}\Url}\fi

\bibitem[{Amazon AGI}(2025)]{amazon2024nova}
{Amazon AGI}.
\newblock The {Amazon Nova} family of models: Technical report and model card.
\newblock \emph{arXiv preprint arXiv:2506.12103}, 2025.
\newblock URL \url{https://arxiv.org/abs/2506.12103}.

\bibitem[{Anthropic}(2026)]{anthropic2026claude}
{Anthropic}.
\newblock {Claude} model overview.
\newblock \url{https://docs.anthropic.com/en/docs/about-claude/models}, 2026.

\bibitem[Canaverde et~al.(2025)Canaverde, Pires, Ribeiro, and
  Martins]{canaverde2025legalbenchpt}
Beatriz Canaverde, Telmo~Pessoa Pires, Leonor~Melo Ribeiro, and Andr{\'e} F.~T.
  Martins.
\newblock {LegalBench.PT}: A benchmark for {Portuguese} law.
\newblock \emph{arXiv preprint arXiv:2502.16357}, 2025.
\newblock URL \url{https://arxiv.org/abs/2502.16357}.

\bibitem[Chalkidis et~al.(2022)Chalkidis, Jana, Hartung, Bommarito,
  Androutsopoulos, Katz, and Aletras]{chalkidis2022lexglue}
Ilias Chalkidis, Abhik Jana, Dirk Hartung, Michael Bommarito, Ion
  Androutsopoulos, Daniel~Martin Katz, and Nikolaos Aletras.
\newblock Lexglue: A benchmark dataset for legal language understanding in
  english.
\newblock In \emph{Proceedings of the 60th Annual Meeting of the Association
  for Computational Linguistics}, pages 4310--4330, 2022.
\newblock URL \url{https://arxiv.org/abs/2110.00976}.

\bibitem[Cohen(1960)]{cohen1960coefficient}
Jacob Cohen.
\newblock A coefficient of agreement for nominal scales.
\newblock \emph{Educational and Psychological Measurement}, 20\penalty0
  (1):\penalty0 37--46, 1960.

\bibitem[{Cour de cassation}(2026)]{src_judilibre}
{Cour de cassation}.
\newblock {Judilibre}: open data of french court decisions.
\newblock \url{https://www.courdecassation.fr/acces-rapide-judilibre}, 2026.

\bibitem[Dietterich(1998)]{dietterich1998approximate}
Thomas~G. Dietterich.
\newblock Approximate statistical tests for comparing supervised classification
  learning algorithms.
\newblock \emph{Neural Computation}, 10\penalty0 (7):\penalty0 1895--1923,
  1998.

\bibitem[Dong et~al.(2025)]{dong2025vlegalbench}
Nguyen~Tien Dong et~al.
\newblock {VLegal-Bench}: Cognitively grounded benchmark for {Vietnamese} legal
  reasoning of large language models.
\newblock \emph{arXiv preprint arXiv:2512.14554}, 2025.
\newblock URL \url{https://arxiv.org/abs/2512.14554}.

\bibitem[Fan et~al.(2025)Fan, Ni, Merane, Tian, Hermstr{\"u}wer, Huang,
  et~al.]{niklaus2025lexam}
Yu~Fan, Jingwei Ni, Jakob Merane, Yang Tian, Yoan Hermstr{\"u}wer, Yinya Huang,
  et~al.
\newblock {LEXam}: Benchmarking legal reasoning on 340 law exams.
\newblock \emph{arXiv preprint arXiv:2505.12864}, 2025.
\newblock URL \url{https://arxiv.org/abs/2505.12864}.

\bibitem[Fujita et~al.(2025)Fujita, Naraki, Zhu, and
  Mori]{fujita2025legalrikai}
Shogo Fujita, Yuji Naraki, Yiqing Zhu, and Shinsuke Mori.
\newblock {LegalRikai}: Open benchmark for complex {Japanese} corporate legal
  tasks.
\newblock \emph{arXiv preprint arXiv:2512.11297}, 2025.
\newblock URL \url{https://arxiv.org/abs/2512.11297}.

\bibitem[Geirhos et~al.(2020)Geirhos, Jacobsen, Michaelis, Zemel, Brendel,
  Bethge, and Wichmann]{geirhos2020shortcut}
Robert Geirhos, J{\"o}rn-Henrik Jacobsen, Claudio Michaelis, Richard Zemel,
  Wieland Brendel, Matthias Bethge, and Felix~A. Wichmann.
\newblock Shortcut learning in deep neural networks.
\newblock \emph{Nature Machine Intelligence}, 2\penalty0 (11):\penalty0
  665--673, 2020.

\bibitem[Grattafiori et~al.(2024)]{grattafiori2024llama3}
Aaron Grattafiori et~al.
\newblock The {Llama} 3 herd of models.
\newblock \emph{arXiv preprint arXiv:2407.21783}, 2024.

\bibitem[Guha et~al.(2023)Guha, Nyarko, Ho, R{\'e}, Chilton, Narayana,
  Chohlas-Wood, Peters, Waldon, Rockmore, et~al.]{guha2024legalbench}
Neel Guha, Julian Nyarko, Daniel~E. Ho, Christopher R{\'e}, Adam Chilton,
  Aditya Narayana, Alex Chohlas-Wood, Austin Peters, Brandon Waldon, Daniel~N.
  Rockmore, et~al.
\newblock Legalbench: A collaboratively built benchmark for measuring legal
  reasoning in large language models.
\newblock In \emph{Advances in Neural Information Processing Systems},
  volume~36, 2023.
\newblock URL \url{https://arxiv.org/abs/2308.11462}.

\bibitem[Gururangan et~al.(2018)Gururangan, Swayamdipta, Levy, Schwartz,
  Bowman, and Smith]{gururangan2018annotation}
Suchin Gururangan, Swabha Swayamdipta, Omer Levy, Roy Schwartz, Samuel~R.
  Bowman, and Noah~A. Smith.
\newblock Annotation artifacts in natural language inference data.
\newblock In \emph{Proceedings of NAACL-HLT}, pages 107--112, 2018.

\bibitem[Hendrycks et~al.(2021)Hendrycks, Burns, Chen, and
  Ball]{hendrycks2021cuad}
Dan Hendrycks, Collin Burns, Anya Chen, and Spencer Ball.
\newblock Cuad: An expert-annotated {NLP} dataset for legal contract review.
\newblock In \emph{Proceedings of the 35th International Conference on Neural
  Information Processing Systems}, 2021.
\newblock URL \url{https://arxiv.org/abs/2103.06268}.

\bibitem[Hijazi et~al.(2024)Hijazi, AlHarbi, AlHussein, Abu~Shairah, AlZahrani,
  AlShamlan, Knio, and Turkiyyah]{hijazi2024arablegaleval}
Faris Hijazi, Somayah AlHarbi, Abdulaziz AlHussein, Harethah Abu~Shairah, Reem
  AlZahrani, Hebah AlShamlan, Omar Knio, and George Turkiyyah.
\newblock {ArabLegalEval}: A multitask benchmark for assessing {Arabic} legal
  knowledge in large language models.
\newblock In \emph{Proceedings of the Second Arabic Natural Language Processing
  Conference (ArabicNLP 2024)}, 2024.
\newblock URL \url{https://arxiv.org/abs/2408.07983}.

\bibitem[Holm(1979)]{holm1979simple}
Sture Holm.
\newblock A simple sequentially rejective multiple test procedure.
\newblock \emph{Scandinavian Journal of Statistics}, 6\penalty0 (2):\penalty0
  65--70, 1979.

\bibitem[Ioannou et~al.(2025)Ioannou, Shiamishis, Hollenstein, and
  G{\"u}rel]{ioannou2025limits}
Antreas Ioannou, Andreas Shiamishis, Nora Hollenstein, and Nezihe~Merve
  G{\"u}rel.
\newblock Evaluating the limits of large language models in multilingual legal
  reasoning.
\newblock \emph{arXiv preprint arXiv:2509.22472}, 2025.
\newblock URL \url{https://arxiv.org/abs/2509.22472}.

\bibitem[McCoy et~al.(2019)McCoy, Pavlick, and Linzen]{mccoy2019right}
R.~Thomas McCoy, Ellie Pavlick, and Tal Linzen.
\newblock Right for the wrong reasons: Diagnosing syntactic heuristics in
  natural language inference.
\newblock In \emph{Proceedings of ACL}, pages 3428--3448, 2019.

\bibitem[McNemar(1947)]{mcnemar1947note}
Quinn McNemar.
\newblock Note on the sampling error of the difference between correlated
  proportions or percentages.
\newblock \emph{Psychometrika}, 12\penalty0 (2):\penalty0 153--157, 1947.

\bibitem[{Meta AI}(2025)]{meta2025llama4}
{Meta AI}.
\newblock The {Llama} 4 herd: The beginning of a new era of natively multimodal
  {AI} innovation.
\newblock \url{https://ai.meta.com/blog/llama-4-multimodal-intelligence/},
  2025.

\bibitem[{Ministry of Justice of Poland} and {Naczelny S{\k{a}}d
  Administracyjny}(2026)]{src_pl}
{Ministry of Justice of Poland} and {Naczelny S{\k{a}}d Administracyjny}.
\newblock {Portal Orzecze{\'n} S{\k{a}}d{\'o}w Powszechnych} and {Centralna
  Baza Orzecze{\'n} S{\k{a}}d{\'o}w Administracyjnych}.
\newblock \url{https://orzeczenia.ms.gov.pl},
  \url{https://orzeczenia.nsa.gov.pl}, 2026.

\bibitem[{Mistral AI}(2025)]{mistral2025large3}
{Mistral AI}.
\newblock Introducing {Mistral 3}.
\newblock \url{https://mistral.ai/news/mistral-3/}, 2025.
\newblock Announcement of Mistral Large 3 and Ministral 3 (3B, 8B, 14B), 2
  December 2025.

\bibitem[{Nacionalin{\.e} teism{\k{u}} administracija}(2026)]{src_liteko}
{Nacionalin{\.e} teism{\k{u}} administracija}.
\newblock {LITEKO}: public search of lithuanian court decisions.
\newblock \url{https://liteko.teismai.lt}, 2026.

\bibitem[Niklaus et~al.(2022)Niklaus, St{\"u}rmer, and
  Chalkidis]{niklaus2022crossx}
Joel Niklaus, Matthias St{\"u}rmer, and Ilias Chalkidis.
\newblock An empirical study on cross-x transfer for legal judgment prediction.
\newblock In \emph{Proceedings of the 2nd Conference of the Asia-Pacific
  Chapter of the Association for Computational Linguistics}, 2022.
\newblock URL \url{https://arxiv.org/abs/2209.12325}.

\bibitem[Niklaus et~al.(2023)Niklaus, Matoshi, Rani, Galassi, St{\"u}rmer, and
  Chalkidis]{niklaus2023lextreme}
Joel Niklaus, Veton Matoshi, Pooja Rani, Andrea Galassi, Matthias St{\"u}rmer,
  and Ilias Chalkidis.
\newblock Lextreme: A multi-lingual and multi-task benchmark for the legal
  domain.
\newblock In \emph{Findings of the Association for Computational Linguistics:
  EMNLP 2023}, pages 4973--5006, 2023.
\newblock URL \url{https://arxiv.org/abs/2301.13126}.

\bibitem[Niklaus et~al.(2024)Niklaus, Matoshi, St{\"u}rmer, Chalkidis, and
  Ho]{niklaus2024multilegal}
Joel Niklaus, Veton Matoshi, Matthias St{\"u}rmer, Ilias Chalkidis, and
  Daniel~E. Ho.
\newblock {MultiLegalPile}: A 689{GB} multilingual legal corpus.
\newblock In \emph{Proceedings of the 62nd Annual Meeting of the Association
  for Computational Linguistics}, 2024.
\newblock URL \url{https://arxiv.org/abs/2306.02069}.

\bibitem[{NVIDIA}(2025)]{nvidia2025nemotronnano2}
{NVIDIA}.
\newblock {NVIDIA Nemotron Nano V2 VL}.
\newblock \emph{arXiv preprint arXiv:2511.03929}, 2025.
\newblock URL \url{https://arxiv.org/abs/2511.03929}.

\bibitem[{NVIDIA}(2026)]{nvidia2026nemotron3super}
{NVIDIA}.
\newblock {Nemotron 3 Super}: Open, efficient mixture-of-experts hybrid
  {Mamba}-transformer model for agentic reasoning.
\newblock \emph{arXiv preprint arXiv:2604.12374}, 2026.
\newblock URL \url{https://arxiv.org/abs/2604.12374}.

\bibitem[Ovcharov(2026{\natexlab{a}})]{ovcharov2025crosslingual}
Volodymyr Ovcharov.
\newblock The tokenizer tax across 25 {European} languages: Domain invariance,
  cross-lingual few-shot effects, and the {Ukrainian} penalty.
\newblock \emph{arXiv preprint arXiv:2605.24718}, 2026{\natexlab{a}}.
\newblock URL \url{https://arxiv.org/abs/2605.24718}.

\bibitem[Ovcharov(2026{\natexlab{b}})]{ovcharov2025ualegalbench}
Volodymyr Ovcharov.
\newblock {UA-Legal-Bench}: A benchmark for evaluating large language models on
  {Ukrainian} legal reasoning.
\newblock \emph{arXiv preprint arXiv:2605.29170}, 2026{\natexlab{b}}.
\newblock URL \url{https://arxiv.org/abs/2605.29170}.

\bibitem[Ovcharov(2026{\natexlab{c}})]{ovcharov2026optionleakage}
Volodymyr Ovcharov.
\newblock Gated against one model, open to the next: Option-only solvability in
  legal multiple-choice benchmarks.
\newblock \emph{arXiv preprint arXiv:2608.15428}, 2026{\natexlab{c}}.

\bibitem[Poliak et~al.(2018)Poliak, Naradowsky, Haldar, Rudinger, and
  Van~Durme]{poliak2018hypothesis}
Adam Poliak, Jason Naradowsky, Aparajita Haldar, Rachel Rudinger, and Benjamin
  Van~Durme.
\newblock Hypothesis only baselines in natural language inference.
\newblock In \emph{Proceedings of the Seventh Joint Conference on Lexical and
  Computational Semantics (*SEM)}, pages 180--191, 2018.

\bibitem[{Raad voor de rechtspraak}(2026)]{src_rechtspraak}
{Raad voor de rechtspraak}.
\newblock Open data van de rechtspraak.
\newblock \url{https://www.rechtspraak.nl/Uitspraken/Paginas/Open-Data.aspx},
  2026.

\bibitem[Rust et~al.(2021)Rust, Pfeiffer, Vuli{\'c}, Ruder, and
  Gurevych]{rust2021tokenizer}
Phillip Rust, Jonas Pfeiffer, Ivan Vuli{\'c}, Sebastian Ruder, and Iryna
  Gurevych.
\newblock How good is your tokenizer? on the monolingual performance of
  multilingual language models.
\newblock In \emph{Proceedings of ACL-IJCNLP}, pages 3118--3135, 2021.

\bibitem[Sainz et~al.(2023)Sainz, Campos, Garc{\'i}a-Ferrero, Etxaniz, Lopez~de
  Lacalle, and Agirre]{sainz2023contamination}
Oscar Sainz, Jon~Ander Campos, Iker Garc{\'i}a-Ferrero, Julen Etxaniz, Oier
  Lopez~de Lacalle, and Eneko Agirre.
\newblock {NLP} evaluation in trouble: On the need to measure {LLM} data
  contamination for each benchmark.
\newblock In \emph{Findings of EMNLP}, 2023.

\bibitem[{State Judicial Administration of Ukraine}(2026)]{src_edrsr}
{State Judicial Administration of Ukraine}.
\newblock {Yedynyi derzhavnyi reiestr sudovykh rishen} ({EDRSR}), unified state
  register of court decisions of ukraine.
\newblock \url{https://reyestr.court.gov.ua}, 2026.

\bibitem[Stern et~al.(2024)Stern, Rasiah, Matoshi, Br{\"u}gger~Bose,
  St{\"u}rmer, Chalkidis, Ho, and Niklaus]{niklaus2024scale}
Ronja Stern, Vishvaksenan Rasiah, Veton Matoshi, Srinanda Br{\"u}gger~Bose,
  Matthias St{\"u}rmer, Ilias Chalkidis, Daniel~E Ho, and Joel Niklaus.
\newblock One law, many languages: Benchmarking multilingual legal reasoning
  for judicial support.
\newblock In \emph{ICLR 2024 Workshop on Data-centric Machine Learning Research
  (DMLR)}, 2024.
\newblock URL \url{https://arxiv.org/abs/2306.09237}.

\bibitem[{{\'U}stavn{\'i} soud} and {Nejvy{\v{s}}{\v{s}}{\'i}
  soud}(2026)]{src_cz}
{{\'U}stavn{\'i} soud} and {Nejvy{\v{s}}{\v{s}}{\'i} soud}.
\newblock {NALUS} (constitutional court) and the decision database of the
  {Nejvy{\v{s}}{\v{s}}{\'i} soud}.
\newblock \url{https://nalus.usoud.cz}, \url{https://sbirka.nsoud.cz}, 2026.

\bibitem[Yang et~al.(2025)]{yang2025qwen3}
An~Yang et~al.
\newblock {Qwen3} technical report.
\newblock \emph{arXiv preprint arXiv:2505.09388}, 2025.

\end{thebibliography}

\clearpage
\appendix
\section{Per-cell data statistics}
\label{app:cellstats}

\begin{table}[H]
\centering
\caption{Per-cell statistics, zero-shot test set (exemplars held out, unanswerable CCP documents removed). \emph{Cls}: offered / realised classes. \emph{Maj}, \emph{Scan}: \% (Section~\ref{sec:leakage}). \emph{Median}, \emph{Mean}: characters of the decision after boilerplate stripping. \emph{Shown}: mean share (\%) of the decision within the input cap. \emph{Labels}: share (\%) of the four most frequent classes.}
\label{tab:cellstats}
\small
\setlength{\tabcolsep}{4pt}
\begin{tabular}{lrrrrrrrp{4.4cm}}
\toprule
\textbf{Cell} & \textbf{$n$} & \textbf{Cls} & \textbf{Maj} & \textbf{Scan} & \textbf{Median} & \textbf{Mean} & \textbf{Shown} & \textbf{Labels} \\
\midrule
FR CTC & 996 & 3/3 & 33 & 84 & 14,014 & 16,701 & 52 & {\scriptsize appel 33, cassation 33, tribunal 33} \\
FR JFC & 787 & 6/6 & 21 & 45 & 6,993 & 11,707 & 71 & {\scriptsize qpc 21, arret 21, autre 21, ordonnance 21, +2 more} \\
FR COP & 810 & 7/7 & 32 & 32 & 825 & 1,751 & 94 & {\scriptsize granted 32, annulled 23, denied 20, no\_ruling 10, +3 more} \\
FR CCP & 928 & 10/10 & 34 & 11 & 21,784 & 23,337 & 36 & {\scriptsize labor 34, contracts 34, business 9, property 9, +6 more} \\
NL CTC & 993 & 6/6 & 17 & 17 & 12,477 & 14,905 & 56 & {\scriptsize cassation 17, first\_multi 17, summary 17, appeal 17, +2 more} \\
NL JFC & 998 & 2/2 & 50 & 96 & 17,677 & 19,921 & 45 & {\scriptsize conclusie 50, uitspraak 50} \\
NL CCP & 996 & 11/3 & 47 & 47 & 11,550 & 14,452 & 57 & {\scriptsize administrative 47, civil 34, criminal 19} \\
PL CTC & 495 & 6/3 & 33 & 33 & 22,345 & 22,719 & 39 & {\scriptsize administrative 33, common 33, ordinary 33} \\
PL JFC & 996 & 3/3 & 33 & 66 & 19,594 & 21,227 & 42 & {\scriptsize postanowienie 33, wyrok 33, wyrok\_uzasadnienie 33} \\
CZ CTC & 332 & 3/1 & 100 & 100 & 10,371 & 11,554 & 62 & {\scriptsize constitutional 100} \\
CZ JFC & 666 & 3/3 & 50 & 56 & 15,619 & 18,585 & 46 & {\scriptsize nalez 50, usneseni 50, stanovisko 0} \\
LT CTC & 997 & 4/4 & 25 & 49 & 15,410 & 17,773 & 49 & {\scriptsize administrative 25, admin\_offense 25, civil 25, criminal 25} \\
LT CCP & 961 & 10/10 & 59 & 59 & 15,620 & 18,360 & 48 & {\scriptsize criminal 59, procedure 19, family 7, civil\_contracts 7, +6 more} \\
 
\bottomrule
\end{tabular}
\end{table}

\begin{table}[H]
\centering
\caption{Full label distribution: every offered class of every cell with its count in the zero-shot test set. Classes with count 0 are offered by the prompt but not realised in the sample.}
\label{tab:labeldist}
\small
\begin{tabular}{lrp{11cm}}
\toprule
\textbf{Cell} & \textbf{$n$} & \textbf{Classes (count)} \\
\midrule
FR CTC & 996 & {\scriptsize appel~332, cassation~332, tribunal~332} \\
FR JFC & 787 & {\scriptsize qpc~166, arret~165, autre~165, ordonnance~165, avis~117, saisie~9} \\
FR COP & 810 & {\scriptsize granted~263, annulled~187, denied~165, no\_ruling~78, inadmissible~46, other~36, forfeiture~35} \\
FR CCP & 928 & {\scriptsize labor~318, contracts~313, business~87, property~80, persons~52, administrative~23, family~23, procedure~12, tort~11, criminal~9} \\
NL CTC & 993 & {\scriptsize cassation~166, first\_multi~166, summary~166, appeal~165, first\_single~165, provisional~165} \\
NL JFC & 998 & {\scriptsize conclusie~499, uitspraak~499} \\
NL CCP & 996 & {\scriptsize administrative~470, civil~334, criminal~192, contracts~0, environmental~0, family~0, immigration~0, insolvency~0, labor~0, social\_security~0, tax~0} \\
PL CTC & 495 & {\scriptsize administrative~165, common~165, ordinary~165, appeal\_chamber~0, constitutional~0, supreme~0} \\
PL JFC & 996 & {\scriptsize postanowienie~332, wyrok~332, wyrok\_uzasadnienie~332} \\
CZ CTC & 332 & {\scriptsize constitutional~332, supreme~0, supreme\_admin~0} \\
CZ JFC & 666 & {\scriptsize nalez~332, usneseni~332, stanovisko~2} \\
LT CTC & 997 & {\scriptsize administrative~250, admin\_offense~249, civil~249, criminal~249} \\
LT CCP & 961 & {\scriptsize criminal~566, procedure~178, family~67, civil\_contracts~64, property~44, administrative~13, insolvency~12, labor~10, tax~4, eu\_international~3} \\
 
\bottomrule
\end{tabular}
\end{table}

\begin{table}[H]
\centering
\caption{Court composition (\%) of cells whose label is not the court itself. \emph{ECLI}: from the court code in the European Case Law Identifier; \emph{registry}: from the \emph{Teismas} field of the LITEKO record; \emph{header}: from the court named in the opening of the decision (Section~\ref{sec:data}). Shares below 1\% omitted.}
\label{tab:courts}
\small
\begin{tabular}{llp{10cm}}
\toprule
\textbf{Cell} & \textbf{Source} & \textbf{Courts} \\
\midrule
FR JFC & header & {\scriptsize Cour de cassation 57, Cour d'appel 36, Tribunal judiciaire 7} \\
FR COP & header & {\scriptsize Cour de cassation 93, Cour d'appel 7} \\
FR CCP & header & {\scriptsize Cour d'appel 89, Cour de cassation 10} \\
NL CTC & ECLI & {\scriptsize Rechtbank 58, Hoge Raad 17, Gerechtshof 10, Raad van State 9, Centrale Raad van Beroep 3, College van Beroep voor het bedrijfsleven 2, Gerecht in eerste aanleg (Caribbean) 1} \\
NL JFC & ECLI & {\scriptsize Parket bij de Hoge Raad (Procurator General) 50, Rechtbank 33, Gerechtshof 7, Raad van State 4, Hoge Raad 2, Centrale Raad van Beroep 2} \\
NL CCP & ECLI & {\scriptsize Rechtbank 68, Gerechtshof 14, Raad van State 7, Hoge Raad 4, Centrale Raad van Beroep 4, Gerecht in eerste aanleg (Caribbean) 2, Parket bij de Hoge Raad (Procurator General) 1} \\
PL CTC & header & {\scriptsize Sąd Okręgowy 33, Wojewódzki Sąd Administracyjny 24, Sąd Rejonowy 19, Naczelny Sąd Administracyjny 10, Sąd Apelacyjny 8, not identified 6} \\
PL JFC & header & {\scriptsize Sąd Okręgowy 46, Sąd Rejonowy 39, Sąd Apelacyjny 15} \\
CZ JFC & ECLI & {\scriptsize Ústavní soud 100} \\
LT CTC & registry & {\scriptsize apylinkės teismas 47, apygardos teismas 21, apygardos administracinis teismas 13, Lietuvos vyriausiasis administracinis teismas 12, Lietuvos Aukščiausiasis Teismas 4, Lietuvos apeliacinis teismas 4} \\
LT CCP & registry & {\scriptsize apylinkės teismas 57, apygardos teismas 27, Lietuvos vyriausiasis administracinis teismas 6, apygardos administracinis teismas 5, Lietuvos apeliacinis teismas 4} \\
 
\bottomrule
\end{tabular}
\end{table}

\begin{table}[H]
\centering
\caption{Court coding scheme. $^{*}$Level read per decision from the case number (Section~\ref{sec:data}).}
\label{tab:scheme}
\footnotesize
\setlength{\tabcolsep}{3pt}
\begin{tabular}{llp{2.3cm}lp{2.6cm}}
\toprule
\textbf{Jur.} & \textbf{Court} & \textbf{Level} & \textbf{Branch} & \textbf{Review} \\
\midrule
FR & Cour de cassation & apex & general & law only \\
FR & Cour d'appel & appeal & general & facts and law \\
FR & Tribunal judiciaire & first instance & general & facts and law \\
FR & Tribunal de commerce & first instance & specialised & facts and law \\
FR & Conseil de prud'hommes & first instance & specialised & facts and law \\
FR & Tribunal (other) & first instance & general & facts and law \\
NL & Rechtbank & first instance & general & facts and law \\
NL & Gerechtshof & appeal & general & facts and law \\
NL & Hoge Raad & apex & general & law only \\
NL & Parket bij de Hoge Raad (Procurator General) & advisory & general & advisory \\
NL & Raad van State & apex & administrative & facts and law \\
NL & Centrale Raad van Beroep & apex & administrative & facts and law \\
NL & College van Beroep voor het bedrijfsleven & apex & administrative & facts and law \\
NL & Gerecht in eerste aanleg (Caribbean) & first instance & general & facts and law \\
NL & Gemeenschappelijk Hof van Justitie (Caribbean) & appeal & general & facts and law \\
NL & Raad van Beroep ambtenarenzaken (Caribbean) & first instance & administrative & facts and law \\
PL & Sąd Rejonowy & first instance & general & facts and law \\
PL & Sąd Okręgowy & first or appeal$^{*}$ & general & facts and law \\
PL & Sąd Apelacyjny & appeal & general & facts and law \\
PL & Sąd Najwyższy & apex & general & law only \\
PL & Wojewódzki Sąd Administracyjny & first instance & administrative & law only \\
PL & Naczelny Sąd Administracyjny & apex & administrative & law only \\
PL & Krajowa Izba Odwoławcza & first instance & specialised & facts and law \\
PL & Trybunał Konstytucyjny & constitutional & constitutional & constitutional review \\
CZ & Ústavní soud & constitutional & constitutional & constitutional review \\
LT & apylinkės teismas & first instance & general & facts and law \\
LT & apygardos teismas & first or appeal$^{*}$ & general & facts and law \\
LT & apygardos administracinis teismas & first instance & administrative & facts and law \\
LT & Lietuvos apeliacinis teismas & appeal & general & facts and law \\
LT & Lietuvos Aukščiausiasis Teismas & apex & general & law only \\
LT & Lietuvos vyriausiasis administracinis teismas & apex & administrative & facts and law \\
 
\bottomrule
\end{tabular}
\end{table}

\begin{table}[H]
\centering
\caption{Record schema of the released data (one JSON object per decision and task).}
\label{tab:schema}
\small
\begin{tabular}{llp{8cm}}
\toprule
\textbf{Field} & \textbf{Type} & \textbf{Content} \\
\midrule
\texttt{doc\_id} & string & Registry identifier (ECLI for NL and CZ, registry id otherwise) \\
\texttt{jurisdiction} & string & \texttt{FR}, \texttt{NL}, \texttt{PL}, \texttt{CZ}, \texttt{LT} \\
\texttt{language} & string & ISO 639-1 code of the decision (\texttt{fr}, \texttt{nl}, \texttt{pl}, \texttt{cs}, \texttt{lt}) \\
\texttt{task} & string & \texttt{ctc}, \texttt{jfc}, \texttt{cop}, \texttt{ccp}, \texttt{ne} \\
\texttt{label\_original} & string (list for NE) & The registry's own value (e.g.\ \emph{Cour d'appel}, \emph{Usnesen\'i}, a subject-area string for CCP); for NE, the cited provisions as published \\
\texttt{label\_normalized} & string (list for NE) & CTC, JFC, COP: the label in the cell's answer space (Table~\ref{tab:labeldist}). CCP: the registry category, mapped to the macro-categories at evaluation time. NE: the cited provisions \\
\texttt{decision\_date} & date & Date of the decision (2020--2026) \\
\texttt{text\_length} & integer & Characters of \texttt{full\_text} \\
\texttt{full\_text} & string & Text of the decision as published by the registry \\
\texttt{facts}, \texttt{facts\_length} & string, integer & COP only: the facts section with the ruling removed, and its length \\
\texttt{norms}, \texttt{norms\_count} & list, integer & NE only: gold citations and their number \\
\bottomrule
\end{tabular}
\end{table}

\begin{table}[H]
\centering
\caption{Masked control per model, zero-shot. $p$: McNemar exact test on the documents whose correctness changes; Holm: adjusted across all 54 model--cell pairs.}
\label{tab:masked_full}
\scriptsize
\setlength{\tabcolsep}{4pt}
\begin{tabular}{llrrrrrr}
\toprule
\textbf{Cell} & \textbf{Model} & \textbf{$n$} & \textbf{Plain} & \textbf{Masked} & \textbf{$\Delta$} & \textbf{$p$} & \textbf{Holm} \\
\midrule
CZ JFC & Maverick & 666 & 75.7 & 50.9 & $-$24.8 & $<$0.001 & $<$0.001 \\
CZ JFC & Llama 3.3 & 666 & 93.1 & 88.3 & $-$4.8 & $<$0.001 & 0.008 \\
CZ JFC & Mistral L3 & 666 & 89.8 & 87.4 & $-$2.4 & 0.11 & 1.00 \\
CZ JFC & Nemotron S3 & 666 & 77.2 & 51.4 & $-$25.8 & $<$0.001 & $<$0.001 \\
CZ JFC & Nova Pro & 666 & 76.0 & 50.8 & $-$25.2 & $<$0.001 & $<$0.001 \\
CZ JFC & Qwen3 235B & 666 & 70.9 & 44.9 & $-$26.0 & $<$0.001 & $<$0.001 \\
CZ JFC & Qwen3 32B & 666 & 69.8 & 54.4 & $-$15.5 & $<$0.001 & $<$0.001 \\
CZ JFC & Sonnet 5 & 666 & 95.5 & 91.3 & $-$4.2 & $<$0.001 & 0.01 \\
CZ JFC & Haiku 4.5 & 666 & 57.8 & 49.8 & $-$8.0 & $<$0.001 & 0.01 \\
FR CTC & Maverick & 996 & 99.5 & 93.5 & $-$6.0 & $<$0.001 & $<$0.001 \\
FR CTC & Llama 3.3 & 996 & 99.7 & 99.5 & $-$0.2 & 0.69 & 1.00 \\
FR CTC & Mistral L3 & 996 & 99.2 & 98.1 & $-$1.1 & 0.05 & 0.68 \\
FR CTC & Nemotron S3 & 996 & 99.3 & 99.1 & $-$0.2 & 0.80 & 1.00 \\
FR CTC & Nova Pro & 996 & 99.8 & 97.4 & $-$2.4 & $<$0.001 & $<$0.001 \\
FR CTC & Qwen3 235B & 996 & 99.9 & 99.1 & $-$0.8 & 0.02 & 0.30 \\
FR CTC & Qwen3 32B & 996 & 99.1 & 92.4 & $-$6.7 & $<$0.001 & $<$0.001 \\
FR CTC & Sonnet 5 & 996 & 99.9 & 99.9 & +0.0 & 1.00 & 1.00 \\
FR CTC & Haiku 4.5 & 996 & 99.2 & 97.2 & $-$2.0 & $<$0.001 & 0.01 \\
FR JFC & Maverick & 787 & 61.4 & 59.1 & $-$2.3 & 0.17 & 1.00 \\
FR JFC & Llama 3.3 & 787 & 54.3 & 31.4 & $-$22.9 & $<$0.001 & $<$0.001 \\
FR JFC & Mistral L3 & 787 & 70.0 & 65.1 & $-$5.0 & 0.002 & 0.03 \\
FR JFC & Nemotron S3 & 787 & 63.5 & 49.7 & $-$13.9 & $<$0.001 & $<$0.001 \\
FR JFC & Nova Pro & 787 & 64.4 & 48.2 & $-$16.3 & $<$0.001 & $<$0.001 \\
FR JFC & Qwen3 235B & 787 & 77.5 & 65.7 & $-$11.8 & $<$0.001 & $<$0.001 \\
FR JFC & Qwen3 32B & 787 & 66.7 & 36.6 & $-$30.1 & $<$0.001 & $<$0.001 \\
FR JFC & Sonnet 5 & 787 & 75.2 & 74.5 & $-$0.8 & 0.45 & 1.00 \\
FR JFC & Haiku 4.5 & 787 & 63.3 & 48.8 & $-$14.5 & $<$0.001 & $<$0.001 \\
LT CTC & Maverick & 997 & 90.5 & 84.2 & $-$6.3 & $<$0.001 & $<$0.001 \\
LT CTC & Llama 3.3 & 997 & 84.1 & 68.2 & $-$15.8 & $<$0.001 & $<$0.001 \\
LT CTC & Mistral L3 & 997 & 98.6 & 88.6 & $-$10.0 & $<$0.001 & $<$0.001 \\
LT CTC & Nemotron S3 & 997 & 77.4 & 50.3 & $-$27.2 & $<$0.001 & $<$0.001 \\
LT CTC & Nova Pro & 997 & 94.8 & 78.8 & $-$15.9 & $<$0.001 & $<$0.001 \\
LT CTC & Qwen3 235B & 997 & 94.6 & 79.7 & $-$14.8 & $<$0.001 & $<$0.001 \\
LT CTC & Qwen3 32B & 997 & 72.9 & 59.5 & $-$13.4 & $<$0.001 & $<$0.001 \\
LT CTC & Sonnet 5 & 997 & 94.8 & 95.7 & +0.9 & 0.24 & 1.00 \\
LT CTC & Haiku 4.5 & 997 & 85.9 & 67.2 & $-$18.7 & $<$0.001 & $<$0.001 \\
NL JFC & Maverick & 998 & 100.0 & 99.5 & $-$0.5 & 0.06 & 0.75 \\
NL JFC & Llama 3.3 & 998 & 100.0 & 95.0 & $-$5.0 & $<$0.001 & $<$0.001 \\
NL JFC & Mistral L3 & 998 & 100.0 & 84.4 & $-$15.6 & $<$0.001 & $<$0.001 \\
NL JFC & Nemotron S3 & 998 & 99.6 & 70.0 & $-$29.6 & $<$0.001 & $<$0.001 \\
NL JFC & Nova Pro & 998 & 99.4 & 99.1 & $-$0.3 & 0.25 & 1.00 \\
NL JFC & Qwen3 235B & 998 & 99.8 & 59.4 & $-$40.4 & $<$0.001 & $<$0.001 \\
NL JFC & Qwen3 32B & 998 & 99.8 & 52.3 & $-$47.5 & $<$0.001 & $<$0.001 \\
NL JFC & Sonnet 5 & 998 & 100.0 & 99.9 & $-$0.1 & 1.00 & 1.00 \\
NL JFC & Haiku 4.5 & 998 & 100.0 & 66.9 & $-$33.1 & $<$0.001 & $<$0.001 \\
PL JFC & Maverick & 996 & 72.4 & 62.7 & $-$9.7 & $<$0.001 & $<$0.001 \\
PL JFC & Llama 3.3 & 996 & 69.4 & 70.6 & +1.2 & 0.49 & 1.00 \\
PL JFC & Mistral L3 & 996 & 71.8 & 62.1 & $-$9.6 & $<$0.001 & $<$0.001 \\
PL JFC & Nemotron S3 & 996 & 73.8 & 62.1 & $-$11.6 & $<$0.001 & $<$0.001 \\
PL JFC & Nova Pro & 996 & 70.2 & 63.9 & $-$6.3 & $<$0.001 & $<$0.001 \\
PL JFC & Qwen3 235B & 996 & 70.7 & 62.3 & $-$8.3 & $<$0.001 & 0.01 \\
PL JFC & Qwen3 32B & 996 & 67.1 & 57.1 & $-$9.9 & $<$0.001 & $<$0.001 \\
PL JFC & Sonnet 5 & 996 & 69.7 & 68.9 & $-$0.8 & 0.42 & 1.00 \\
PL JFC & Haiku 4.5 & 996 & 71.6 & 66.0 & $-$5.6 & 0.010 & 0.15 \\
 
\bottomrule
\end{tabular}
\end{table}

\begin{table}[H]
\centering
\caption{Answers not given as a single line (\% of the zero-shot test set). Responses are read from their first line, the same rule for every model.}
\label{tab:format}
\scriptsize
\setlength{\tabcolsep}{3pt}
\resizebox{\linewidth}{!}{%
\begin{tabular}{lrrrrrrrrrrrr}
\toprule
\textbf{Model} & FR~CTC & FR~JFC & FR~COP & FR~CCP & NL~CTC & NL~JFC & NL~CCP & PL~CTC & PL~JFC & CZ~JFC & LT~CTC & LT~CCP \\
\midrule
Sonnet 5 & 0 & 1 & 0 & 0 & 2 & 0 & 0 & 1 & 0 & 0 & 0 & 0 \\
Haiku 4.5 & 0 & 0 & 9 & 1 & 11 & 0 & 2 & 7 & 0 & 56 & 9 & 8 \\
 
\bottomrule
\end{tabular}}
\end{table}

\clearpage
\section{Full results}
\label{app:full}

\begin{sidewaystable}
\centering
\caption{Accuracy (\%) of every model in every informative cell, zero-shot (ZS) and 3-shot (FS; Claude was run zero-shot only). Norm extraction is in Table~\ref{tab:ne}.}
\label{tab:full_acc}
\scriptsize
\setlength{\tabcolsep}{3pt}
\begin{tabular}{llrrrrrrrrrrrr}
\toprule
\textbf{Model} & & FR~CTC & FR~JFC & FR~COP & FR~CCP & NL~CTC & NL~JFC & NL~CCP & PL~CTC & PL~JFC & CZ~JFC & LT~CTC & LT~CCP \\
\midrule
Maverick & ZS & 99.5 & 61.4 & 29.5 & 62.4 & 94.0 & 100.0 & 28.8 & 60.0 & 72.4 & 75.7 & 90.5 & 61.3 \\
Llama 3.3 & ZS & 99.7 & 54.3 & 40.0 & 60.8 & 87.0 & 100.0 & 51.3 & 58.8 & 69.4 & 93.1 & 84.1 & 57.8 \\
Mistral L3 & ZS & 99.2 & 70.0 & 40.1 & 60.2 & 97.1 & 100.0 & 43.6 & 55.2 & 71.8 & 89.8 & 98.6 & 61.8 \\
Nemotron S3 & ZS & 99.3 & 63.5 & 39.9 & 67.1 & 82.1 & 99.6 & 63.4 & 59.6 & 73.8 & 77.2 & 77.4 & 53.0 \\
Nova Pro & ZS & 99.8 & 64.4 & 31.9 & 65.6 & 94.6 & 99.4 & 51.8 & 53.1 & 70.2 & 76.0 & 94.8 & 57.6 \\
Qwen3 235B & ZS & 99.9 & 77.5 & 41.0 & 65.1 & 95.0 & 99.8 & 49.7 & 59.2 & 70.7 & 70.9 & 94.6 & 56.1 \\
Qwen3 32B & ZS & 99.1 & 66.7 & 39.1 & 56.1 & 96.7 & 99.8 & 65.2 & 52.3 & 67.1 & 69.8 & 72.9 & 57.3 \\
Sonnet 5 & ZS & 99.9 & 75.2 & 44.2 & 63.9 & 94.7 & 100.0 & 47.0 & 53.7 & 69.7 & 95.5 & 94.8 & 60.4 \\
Haiku 4.5 & ZS & 99.2 & 63.3 & 27.3 & 55.4 & 91.9 & 100.0 & 56.5 & 47.1 & 71.6 & 57.8 & 85.9 & 54.1 \\
\midrule
Maverick & FS & 99.6 & 65.1 & 34.8 & 63.1 & 97.9 & 92.4 & 70.6 & 54.1 & 71.7 & 76.6 & 82.3 & 60.6 \\
Llama 3.3 & FS & 99.6 & 68.7 & 36.8 & 64.8 & 96.3 & 100.0 & 74.7 & 45.3 & 66.5 & 86.9 & 96.0 & 59.2 \\
Mistral L3 & FS & 100.0 & 78.4 & 39.1 & 57.3 & 98.7 & 100.0 & 37.7 & 46.5 & 66.8 & 89.2 & 97.9 & 59.2 \\
Nemotron S3 & FS & 99.5 & 69.8 & 33.8 & 67.0 & 93.4 & 99.4 & 77.1 & 68.5 & 68.8 & 83.2 & 89.2 & 56.0 \\
Nova Pro & FS & 100.0 & 76.7 & 41.7 & 64.2 & 98.9 & 99.5 & 67.7 & 55.2 & 67.4 & 65.2 & 96.7 & 58.8 \\
Qwen3 235B & FS & 99.9 & 77.8 & 40.2 & 63.7 & 98.1 & 100.0 & 62.6 & 56.0 & 64.2 & 66.4 & 95.2 & 60.0 \\
Qwen3 32B & FS & 99.7 & 71.0 & 35.8 & 60.2 & 98.6 & 99.9 & 75.5 & 51.9 & 74.2 & 71.6 & 92.3 & 59.5 \\
Sonnet 5 & FS & -- & -- & -- & -- & -- & -- & -- & -- & -- & -- & -- & -- \\
Haiku 4.5 & FS & -- & -- & -- & -- & -- & -- & -- & -- & -- & -- & -- & -- \\
 
\bottomrule
\end{tabular}

\vspace{1.5em}
\caption{Cohen's $\kappa$ against the gold label, zero-shot. An unmatched response counts as its own category.}
\label{tab:full_kappa}
\begin{tabular}{lrrrrrrrrrrrr}
\toprule
\textbf{Model} & FR~CTC & FR~JFC & FR~COP & FR~CCP & NL~CTC & NL~JFC & NL~CCP & PL~CTC & PL~JFC & CZ~JFC & LT~CTC & LT~CCP \\
\midrule
Maverick & 0.99 & 0.52 & 0.17 & 0.52 & 0.93 & 1.00 & 0.21 & 0.43 & 0.59 & 0.52 & 0.87 & 0.47 \\
Llama 3.3 & 1.00 & 0.43 & 0.21 & 0.51 & 0.84 & 1.00 & 0.41 & 0.42 & 0.54 & 0.86 & 0.79 & 0.43 \\
Mistral L3 & 0.99 & 0.63 & 0.21 & 0.50 & 0.96 & 1.00 & 0.34 & 0.38 & 0.58 & 0.80 & 0.98 & 0.47 \\
Nemotron S3 & 0.99 & 0.55 & 0.23 & 0.56 & 0.78 & 0.99 & 0.52 & 0.42 & 0.61 & 0.55 & 0.70 & 0.36 \\
Nova Pro & 1.00 & 0.56 & 0.15 & 0.55 & 0.93 & 0.99 & 0.41 & 0.37 & 0.55 & 0.52 & 0.93 & 0.43 \\
Qwen3 235B & 1.00 & 0.72 & 0.25 & 0.55 & 0.94 & 1.00 & 0.39 & 0.43 & 0.56 & 0.46 & 0.93 & 0.40 \\
Qwen3 32B & 0.99 & 0.59 & 0.19 & 0.46 & 0.96 & 1.00 & 0.55 & 0.36 & 0.51 & 0.41 & 0.64 & 0.42 \\
Sonnet 5 & 1.00 & 0.69 & 0.31 & 0.55 & 0.94 & 1.00 & 0.37 & 0.35 & 0.55 & 0.91 & 0.93 & 0.46 \\
Haiku 4.5 & 0.99 & 0.54 & 0.13 & 0.46 & 0.90 & 1.00 & 0.46 & 0.33 & 0.57 & 0.36 & 0.81 & 0.38 \\
 
\bottomrule
\end{tabular}
\end{sidewaystable}

\begin{table}[H]
\centering
\caption{Paired comparison, 3-shot. Columns as in Table~\ref{tab:paired}.}
\label{tab:paired_fs}
\small
\setlength{\tabcolsep}{4pt}
\begin{tabular}{lrllrrrcr}
\toprule
\textbf{Cell} & \textbf{$n$} & \textbf{Top} & \textbf{Second} & \textbf{Gap} & \textbf{Disc.} & \textbf{$p$} & \textbf{Leader} & \textbf{Sig.\ pairs} \\
\midrule
FR CTC & 996 & Mistral L3 & Nova Pro & 0.0 & 0/0 & 1.00 & \xmark & 0/21 \\
FR JFC & 787 & Mistral L3 & Qwen3 235B & 0.6 & 5/0 & 0.06 & \xmark & 15/21 \\
FR COP & 810 & Nova Pro & Qwen3 235B & 1.5 & 113/101 & 0.45 & \xmark & 4/21 \\
FR CCP & 928 & Nemotron S3 & Llama 3.3 & 2.3 & 51/30 & 0.03 & \xmark & 8/21 \\
NL CTC & 993 & Nova Pro & Mistral L3 & 0.2 & 4/2 & 0.69 & \xmark & 10/21 \\
NL JFC & 998 & Llama 3.3 & Mistral L3 & 0.0 & 0/0 & 1.00 & \xmark & 6/21 \\
NL CCP & 996 & Nemotron S3 & Qwen3 32B & 1.6 & 48/32 & 0.09 & \xmark & 19/21 \\
PL CTC & 495 & Nemotron S3 & Qwen3 235B & 12.5 & 88/26 & $<$0.001 & \cmark & 14/21 \\
PL JFC & 996 & Qwen3 32B & Maverick & 2.5 & 26/1 & $<$0.001 & \cmark & 17/21 \\
CZ JFC & 666 & Mistral L3 & Llama 3.3 & 2.3 & 59/44 & 0.17 & \xmark & 18/21 \\
LT CTC & 997 & Mistral L3 & Nova Pro & 1.2 & 21/9 & 0.04 & \xmark & 17/21 \\
LT CCP & 961 & Maverick & Qwen3 235B & 0.5 & 31/26 & 0.60 & \xmark & 5/21 \\
 
\bottomrule
\end{tabular}
\end{table}

\begin{table}[H]
\centering
\caption{Norm extraction: mean per-document set F1$\times$100 under normalised (article, law) pair matching, per model. Exact string matching against the registry's reference strings gives 1.1--4.7 (PL) and 0.2--34.6 (CZ); the matching rule moves the score by more than the choice of model. Claude was not run on NE.}
\label{tab:ne}
\small
\begin{tabular}{lrrrr}
\toprule
& \multicolumn{2}{c}{\textbf{PL NE}} & \multicolumn{2}{c}{\textbf{CZ NE}} \\
\cmidrule(lr){2-3} \cmidrule(lr){4-5}
\textbf{Model} & ZS & FS & ZS & FS \\
\midrule
Maverick & 14.0 & 14.1 & 1.4 & 49.6 \\
Llama 3.3 & 14.7 & 13.8 & 1.5 & 5.3 \\
Mistral L3 & 12.4 & 12.7 & 0.4 & 29.0 \\
Nemotron S3 & 12.0 & 14.4 & 1.1 & 22.6 \\
Nova Pro & 12.6 & 12.8 & 0.9 & 18.5 \\
Qwen3 235B & 12.4 & 12.6 & 1.6 & 26.5 \\
Qwen3 32B & 13.1 & 12.9 & 1.2 & 23.2 \\
 
\bottomrule
\end{tabular}
\end{table}

\begin{figure}[H]
\centering
\input{figures/fig_kappa_heatmap.tex}
\caption{Zero-shot Cohen's $\kappa$ of all nine models per informative cell (CZ~CTC omitted), with the label names left in the input. $^\dagger$~Cells whose label names are printed in the decision (Section~\ref{sec:leakage}); for most models part of their $\kappa$ comes from those names; Figure~\ref{fig:masked} gives accuracy with the names masked. Columns ordered by model family.}
\label{fig:model_rankings}
\end{figure}

\section{Scaling and fertility}
\label{app:scaling}

\begin{table}[H]
\centering
\caption{Mean zero-shot accuracy of four small models (3--12B) minus that of the seven Bedrock models, pp.}
\label{tab:scaling}
\small
\begin{tabular}{l R{1.0cm} R{1.0cm} R{1.0cm} R{1.0cm} R{1.0cm}}
\toprule
\textbf{Task} & \textbf{FR} & \textbf{NL} & \textbf{PL} & \textbf{CZ} & \textbf{LT} \\
\midrule
CTC & $-$14 & $-$17 & $-$12 & $-$15 & $-$21 \\
JFC & $-$8  & $-$6  & $-$4  & $-$21 & -- \\
COP & $-$3  & --    & --    & --    & -- \\
CCP & $-$11 & $+$5  & --    & --    & $-$11 \\
\bottomrule
\end{tabular}
\end{table}

\begin{figure}[H]
\centering
\input{figures/fig_fertility_heatmap.tex}
\caption{Tokenizer fertility by language and model, each measured with the model's own tokenizer. Rows with identical values share a tokenizer (Tekken: Mistral Large 3, Ministral 3, Nemotron Super 3, Nemotron Nano 12B; Qwen3). The nine rows are the models with a public tokenizer, including the scaling models Ministral 3B/8B and Nemotron Nano 12B; Nova Pro and the two Claude models are omitted because their tokenizers are not public.}
\label{fig:fertility_heatmap}
\end{figure}

\section{Prompts}
\label{app:prompts}

Zero-shot prompt of every classification cell, verbatim; \texttt{\{text\}} is replaced by the decision. The 3-shot prompt adds three exemplars (the first 500 characters of a held-out decision and its label) under a native-language heading before the decision; one 3-shot prompt is given in full at the end. Ukrainian prompts are given in the companion paper~\citep{ovcharov2025ualegalbench}.

\paragraph{FR CTC}
\begin{small}\begin{verbatim}
[system] Vous etes un expert en classification des decisions judiciaires francaises.
Determinez le type de juridiction de cette decision judiciaire.
Options : cassation, appel, tribunal.
Repondez en UN SEUL MOT -- le type de juridiction.

Texte de la decision :
{text}
\end{verbatim}\end{small}

\paragraph{FR JFC}
\begin{small}\begin{verbatim}
[system] Vous etes un expert en classification des decisions judiciaires francaises.
Determinez le type de cette decision judiciaire.
Options : arret, autre, ordonnance, avis, qpc, saisie.
Repondez en UN SEUL MOT -- le type de decision.

Texte de la decision :
{text}
\end{verbatim}\end{small}

\paragraph{FR COP}
\begin{small}\begin{verbatim}
[system] Vous etes un expert en prediction des resultats judiciaires.
Sur la base des faits de cette affaire, predisez le resultat.
Options : denied (rejet), granted (cassation), inadmissible (irrecevabilite), other
    (autre), forfeiture (decheance), no_ruling (non-lieu), annulled (annulation).
Repondez en UN SEUL MOT (en anglais) -- le resultat.

Faits de l'affaire :
{text}
\end{verbatim}\end{small}

\paragraph{FR CCP}
\begin{small}\begin{verbatim}
[system] Vous etes un expert en classification thematique des decisions judiciaires.
Determinez la categorie thematique de cette decision judiciaire.
Options : labor, contracts, family, criminal, property, business, administrative,
    procedure, tort, persons.
Repondez en UN SEUL MOT (en anglais) -- la categorie.

Texte de la decision :
{text}
\end{verbatim}\end{small}

\paragraph{NL CTC}
\begin{small}\begin{verbatim}
[system] U bent een expert in de classificatie van Nederlandse rechtspraak.
Bepaal het type procedure van deze rechterlijke uitspraak.
Opties: appeal, first_single, first_multi, cassation, provisional, summary.
Antwoord met EEN WOORD (in het Engels) -- het type procedure.

Tekst van de uitspraak:
{text}
\end{verbatim}\end{small}

\paragraph{NL JFC}
\begin{small}\begin{verbatim}
[system] U bent een expert in de classificatie van Nederlandse rechtspraak.
Bepaal het type van dit document.
Opties: uitspraak, conclusie.
Antwoord met EEN WOORD -- het documenttype.

Tekst:
{text}
\end{verbatim}\end{small}

\paragraph{NL CCP}
\begin{small}\begin{verbatim}
[system] U bent een expert in de classificatie van Nederlandse rechtspraak.
Bepaal het rechtsgebied van deze uitspraak.
Opties: civil, criminal, administrative, tax, social_security, immigration, family,
    contracts, environmental, labor, insolvency.
Antwoord met EEN WOORD (in het Engels) -- het rechtsgebied.

Tekst van de uitspraak:
{text}
\end{verbatim}\end{small}

\paragraph{PL CTC}
\begin{small}\begin{verbatim}
[system] Jestes ekspertem w klasyfikacji polskich orzeczen sadowych.
Okresl typ sadu wydajacego to orzeczenie.
Opcje: administrative, ordinary, common, supreme, appeal_chamber, constitutional.
Odpowiedz JEDNYM SLOWEM (po angielsku) -- typ sadu.

Tekst orzeczenia:
{text}
\end{verbatim}\end{small}

\paragraph{PL JFC}
\begin{small}\begin{verbatim}
[system] Jestes ekspertem w klasyfikacji polskich orzeczen sadowych.
Okresl forme tego orzeczenia sadowego.
Opcje: wyrok, wyrok_uzasadnienie, postanowienie.
Odpowiedz JEDNYM SLOWEM -- forma orzeczenia.

Tekst orzeczenia:
{text}
\end{verbatim}\end{small}

\paragraph{CZ CTC}
\begin{small}\begin{verbatim}
[system] Jste expert na klasifikaci ceskych soudnich rozhodnuti.
Urcete typ soudu, ktery vydal toto rozhodnuti.
Moznosti: constitutional, supreme, supreme_admin.
Odpovezte JEDNIM SLOVEM (anglicky) -- typ soudu.

Text rozhodnuti:
{text}
\end{verbatim}\end{small}

\paragraph{CZ JFC}
\begin{small}\begin{verbatim}
[system] Jste expert na klasifikaci ceskych soudnich rozhodnuti.
Urcete formu tohoto soudniho rozhodnuti.
Moznosti: usneseni, nalez, stanovisko.
Odpovezte JEDNIM SLOVEM -- forma rozhodnuti.

Text rozhodnuti:
{text}
\end{verbatim}\end{small}

\paragraph{LT CTC}
\begin{small}\begin{verbatim}
[system] Jus esate Lietuvos teismo sprendimu klasifikavimo ekspertas.
Nustatykite sios bylos tipa.
Pasirinkimai: criminal, civil, admin_offense, administrative.
Atsakykite VIENU ZODZIU (angliskai) -- bylos tipas.

Sprendimo tekstas:
{text}
\end{verbatim}\end{small}

\paragraph{LT CCP}
\begin{small}\begin{verbatim}
[system] Jus esate Lietuvos teismo sprendimu klasifikavimo ekspertas.
Nustatykite sio teismo sprendimo tematine kategorija.
Pasirinkimai: criminal, civil_contracts, family, administrative, labor, property,
    procedure, tax, insolvency, eu_international.
Atsakykite VIENU ZODZIU (angliskai) -- kategorija.

Sprendimo tekstas:
{text}
\end{verbatim}\end{small}

\paragraph{FR CTC, 3-shot (complete)}
\begin{small}\begin{verbatim}
Determinez le type de juridiction de cette decision judiciaire.
Options : cassation, appel, tribunal.
Repondez en UN SEUL MOT -- le type de juridiction.
Exemples :
Exemple 1:
Texte: COUR D’APPEL D’AIX EN PROVENCE
TRIBUNAL JUDICIAIRE DE NICE
GREFFE
(Décision Civile)
JUGEMENT : S.A.R.L. GROUPE FUSSIGNY CONSTRUCTION c/ Société STELLA
MINUTE N°
Du 06 Septembre 2024
2ème Chambre civile
N° RG 24/01574 - N° Portalis DBWR-W-B7I-PV5H
Grosse délivrée à
expédition délivrée à
Me Vanessa HAURET
Me Adrien VERRIER
le 6 Septembre 2024
mentions diverses
-Expertise
-Renvoi MEE 5 décembre 2024
Par jugement de la 2ème Chambre civile
Reponse: tribunal
Exemple 2:
Texte: CIV. 2
CM
COUR DE CASSATION
 ______________________
Audience publique du 25 novembre 2021
Cassation partielle
M. PIREYRE, président
Arrêt n° 1061 FS-B+R
Pourvoi n° W 20-17.234
R É P U B L I Q U E F R A N Ç A I S E
_________________________
AU NOM DU PEUPLE FRANÇAIS
 _________________________
ARRÊT DE LA COUR DE CASSATION, DEUXIÈME CHAMBRE CIVILE, DU 25 NOVEMBRE 2021
M. [W] [V], domicilié [Adresse 1], a formé le pourvoi n° W 20-17.23
Reponse: cassation
Exemple 3:
Texte: Grosse + copie
délivrées le
à
COUR D'APPEL DE MONTPELLIER
3e chambre civile
ARRET DU 18 JANVIER 2024
Numéro d'inscription au répertoire général :
N° RG 19/04809 - N° Portalis DBVK-V-B7D-OHW6
Décision déférée à la Cour :
Jugement du 27 JUIN 2019
TRIBUNAL DE GRANDE INSTANCE DE BEZIERS
N° RG 15/00225
APPELANTE :
Madame [G] [K]
née le 02 Juin 1969 à [Localité 11]
de nationalité Française
[Adresse 6]
[Localité 5]
Représentée par Me Philippe
Reponse: appel
Texte de la decision :
{text}
\end{verbatim}\end{small}

\end{document}